\documentclass{article}

\usepackage[T1]{fontenc}
\usepackage[utf8]{inputenc}
\usepackage{microtype}
\usepackage{graphicx}
\usepackage{float}
\usepackage{subcaption}
\usepackage{booktabs} 
\usepackage{hyperref}
\usepackage{amsfonts}
\usepackage{amsmath}

\usepackage[accepted]{preprint}
\icmltitlerunning{Molecule Attention Transformer}

\begin{document}

\twocolumn[
\icmltitle{Molecule Attention Transformer}

\icmlsetsymbol{equal}{*}

\begin{icmlauthorlist}
\icmlauthor{Łukasz Maziarka}{ardi,uj}
\icmlauthor{Tomasz Danel}{ardi,uj}
\icmlauthor{Sławomir Mucha}{uj}
\icmlauthor{Krzysztof Rataj}{ardi}
\icmlauthor{Jacek Tabor}{uj}
\icmlauthor{Stanisław Jastrzębski}{mol,nyu}
\end{icmlauthorlist}

\icmlaffiliation{ardi}{Ardigen, Cracow, Poland}
\icmlaffiliation{uj}{Jagiellonian University, Cracow, Poland}
\icmlaffiliation{mol}{Molecule.one, Warsaw, Poland}
\icmlaffiliation{nyu}{New York University, New York, USA}

\icmlcorrespondingauthor{Łukasz Maziarka}{lukasz.maziarka@ardigen.com}
\icmlcorrespondingauthor{Stanisław Jastrzębski}{staszek.jastrzebski@gmail.com}

\icmlkeywords{Self-attention, Graph Convolutional Neural Networks, Molecular property prediction}

\vskip 0.3in
]

\printAffiliationsAndNotice{}  

\begin{abstract}
Designing a single neural network architecture that performs competitively across a range of molecule property prediction tasks remains largely an open challenge, and its solution may unlock a widespread use of deep learning in the drug discovery industry. To move towards this goal, we propose Molecule Attention Transformer (MAT). Our key innovation is to augment the attention mechanism in Transformer using inter-atomic distances and the molecular graph structure. Experiments show that MAT performs competitively on a diverse set of molecular prediction tasks. Most importantly, with a simple self-supervised pretraining, MAT requires tuning of only a few hyperparameter values to achieve state-of-the-art performance on downstream tasks. Finally, we show that attention weights learned by MAT are interpretable from the chemical point of view.

\end{abstract}

\section{Introduction}
\label{introduction}

The task of predicting properties of a molecule lies at the center of applications such as drug discovery or material design. In particular, estimated $85\%$ drug candidates fail the clinical trials in the United States after a long and costly development process~\citep{Wong2018}. Potentially, many of these failures could have been avoided by having correctly predicted a clinically relevant property of a molecule such as its toxicity or bioactivity.

Following the breakthroughs in image~\citep{Krizhevsky2012} and text classification~\citep{vaswani2017}, deep neural networks (DNNs) are expected to revolutionize other fields such as drug discovery or material design~\citep{jr2019survey}. However, on many molecular property prediction tasks DNNs are outperformed by \emph{shallow} models such as support vector machine or random forest~\citep{Korotcov2017,Wu2018}. On the other hand, while DNNs can outperform shallow models on some tasks, they tend to be difficult to train~\citep{Ishiguro2019,Hu2019}, and can require tuning of a large number of hyperparameters.  We also observe both issues on our benchmark (see Section \ref{sec:comp_models}).

Making deep networks easier to train has been the central force behind their widespread use. In particular, one of the most important breakthroughs in deep learning was the development of initialization methods that allowed to train easily deep networks end-to-end~\citep{Goodfellow2016}. In a similar spirit, our aim is to develop a deep model that is simple to use out-of-the-box, and achieves strong performance on a wide range of tasks in the field of molecule property prediction.

In this paper we propose the Molecule Attention Transformer (MAT). We adapt Transformer~\citep{devlin2018pretraining} to chemical molecules by augmenting the self-attention with inter-atomic distances and molecular graph structure. Figure~\ref{fig:architecture} shows the architecture. We demonstrate that MAT, in contrast to other tested models, achieves strong performance across a wide range of tasks  (see Figure~\ref{fig:main_results}). Next, we show that self-supervised pre-training further improves performance, while drastically reducing the time needed for hyperparameter tuning (see Table~\ref{fig:results_500_pretrained}). In these experiments we tuned only the learning rate, testing $7$ different values. Finally, we find that MAT has interpretable attention weights. We share pretrained weights at \href{https://github.com/gmum/MAT}{https://github.com/gmum/MAT}.

\section{Related work}
\label{sec:rel_work}

\begin{figure*}[h]
    \centering
    \includegraphics[width=0.62\textwidth]{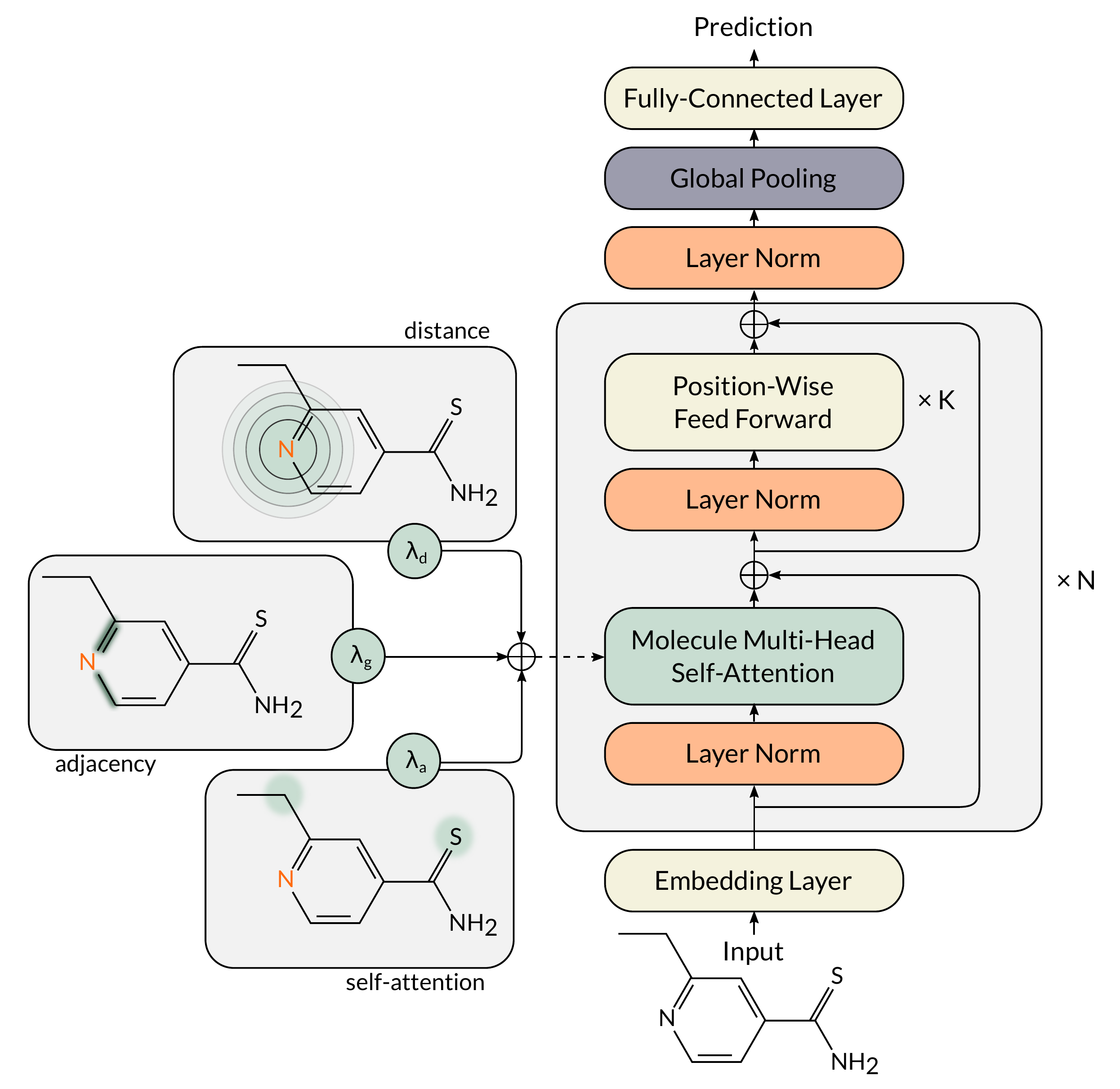}
    \caption{Molecule Attention Transformer architecture. We largely base our model on the Transformer encoder. In the first layer we embed each atom using one-hot encoding and atomic features. The main innovation is the Molecule Multi-Head Self-Attention layer that augments attention with distance and graph structure of the molecule. We implement this using a weighted (by $\lambda_d$, $\lambda_g$, and $\lambda_a$) element-wise sum of the corresponding matrices.}

    \label{fig:architecture}
\end{figure*}

\paragraph{Molecule property prediction.} 

Predicting properties of a candidate molecule lies at the heart of many fields such as drug discovery and material design. Broadly speaking, there are two main approaches to predicting molecular properties. First, we can use our knowledge of the underlying physics~\citep{LIPINSKI19973}. However, despite recent advances~\citep{Schutt2017}, current approaches remain prohibitively costly to accurately predict many properties of interest such as bioactivity. The second approach is to use existing data to train a predictive model~\citep{HAGHIGHATLARI201951}. Here the key issue is the lack of large datasets. Even for the most popular drug targets, such as 5-HT1A (a popular target for depression), only thousands of active compounds are known. Promising direction is using hybrid approaches such as \citet{wallach2015} or approaches leveraging domain knowledge and underlying physics to impose a strong prior such as \citet{Feinberg2018}.

\paragraph{Deep learning for molecule property prediction.} 

Deep learning has become a valuable tool for modeling molecules. During the years, the community has progressed from using handcrafted representations to representing molecules as strings of symbols, and finally to the currently popular approaches based on molecular graphs.

Graph convolutional networks in each subsequent layer gather information from adjacent nodes in the graph. In this way after $N$ convolution layers each node has information from its $N$-edges distant neighbors. Using the graph structure improves performance in a range of molecule modeling tasks~\citep{Wu2018}. Some of the most recent works implement more sophisticated generalization methods for gathering the neighbor data.
\citet{velikovic2017,shang2018} propose to augment GCNs with an attention mechanism. \citet{li2018} introduces a model that dynamically learns neighbourhood function in the graph. 

In parallel to these advances, using the three-dimensional structure of the molecule is becoming increasingly popular. Perhaps the most closely related models are 3D Graph Convolutional Neural Network (3DGCN), Message Passing Neural Network (MPNN), and Adaptive Graph Convolutional Network (AGCN)~\citep{cho2018, gilmer2017, li2018}. 3DGCN and MPNN integrate graph and distance information in a single model, which enables them to achieve strong performance on tasks such as solubility prediction. In contrast to them, we additionally allow for a flexible neighbourhood based on self-attention.

Transformer, originally developed for natural language processing~\citep{vaswani2017}, has been recently applied to retrosynthesis in \citet{karpov2019transformer}. They represent compounds as sentences using the SMILES notation~\citep{weininger1988smiles}. In contrast to them, we represent compounds as a list of atoms, and ensure that models understand the structure of the molecule by augmenting the self-attention mechanism (see Figure~\ref{fig:architecture}). Our ablation studies show it is a critical component of the model. 

To summarize, methods related to our model have been proposed in the literature. Our contribution is unifying these ideas in a single model based on the state-of-the-art Transformer architecture that preserves strong performance across many chemical tasks.

\paragraph{How easy is it to use deep learning for molecule property prediction?}  

DNNs performance is not always competitive to methods such as support vector machine or random forest. MoleculeNet is a popular benchmark for methods for molecule property prediction~\citep{Wu2018} that demonstrates this phenomenon. Similar results can be found in \citet{withnall2019}. We reproduce a similar issue on our benchmark. This makes using deep learning less applicable to molecule property prediction because in some cases practitioners might actually benefit from using other methods. Another issue is that graph neural networks, which are the most popular class of models for molecule property prediction, can be difficult to train. \citet{Ishiguro2019} show and try to address the problem that graph neural networks tend to underfit the training set. We also reproduce this issue on our benchmark (see also App.~\ref{app:more_experiments}).

There has been a considerable interest in developing easier to use deep models for molecule property prediction. \citet{Goh2017chemception} pretrains a deep network that takes as an input an image of a molecule.  Another studies highlight the need to augment feedforward~\citep{Mayr2018} and graph neural networks~\citep{Yang2019} with handcrafted representations of molecules. \citet{Hu2019} proposes pretraining methods for graph neural networks and shows this largely alleviates the problem of underfitting, present in these architectures~\citep{Ishiguro2019}. We take inspiration from \citet{Hu2019} and use one of the three pretraining tasks proposed therein.

Concurrently, \citet{wang2019,honda2019smiles} pretrain a vanilla Transformer~\citep{devlin2018pretraining} that takes as input a text representation (SMILES) of a molecule. \citet{honda2019smiles} shows that decoding based approach improves data efficiency of the model. A similar approach, specialized to the task of drug-target interaction prediction, was concurrently proposed in \citet{shin2019}. In contrast to them, we adapt Transformer to chemical structures, which in our opinion is crucial for achieving strong empirical performance. We also use a domain-specific pretraining based on \citet{Wu2018}. We further confirm importance of both approaches by comparing directly with \citet{honda2019smiles}.

\paragraph{Self-attention based models.} 

Arguably, the attention mechanism~\citep{bahdanau2014neural} has been one of the most important breakthroughs in deep learning. This is perhaps best illustrated by the wide-spread use of Transformer architecture in natural language processing~\citep{vaswani2017,devlin2018pretraining}.

Multiple prior works have augmented self-attention in Transformer using domain-specific knowledge~\citep{Chen2018, shaw2018, bello2019, Guo2019GaussianTA}. \citet{Guo2019GaussianTA} encourages Transformer to attend to adjacent words in a sentence, and \citet{Chen2018} encourages another attention-based model to focus on pairs of words in a sentence that are connected in an external knowledge base. Our novelty is applying this successive modeling idea to molecule property prediction.

\section{Molecule Attention Transformer}
\label{sec:architecture}

As the rich literature on deep learning for molecule property prediction suggests, it is necessary for a model to be flexible enough to represent a range of possible relationships between atoms of a compound. Inspired by its flexibility and strong empirical performance, we base our model on the Transformer encoder~\citep{vaswani2017,devlin2018pretraining}. It is worth noting that natural language processing has inspired important advances in cheminformatics~\citep{Segler2017,Bombarelli2018}, which might be due to similarities between the two domains~\citep{jastrzebski2016}.

\paragraph{Transfomer.}

We begin by briefly introducing the Transformer architecture. On a high level, Transformer for classifications has $N$ attention blocks followed by a pooling and a classification layer. Each attention block is composed of a multi-head self-attention layer, followed by a feed-forward block that includes a residual connection and layer normalization. 

The multi-head self-attention is composed of $H$ heads. Head $i$ ($i=1,\ldots,H$) takes as input hidden state $\mathbf{H}$ and computes first $\mathbf{Q}_i=\mathbf{H}\mathbf{W}_i^Q$, $\mathbf{K}_i=\mathbf{H} \mathbf{W}_i^H$, and $\mathbf{V}_i=\mathbf{H}\mathbf{W}_i^V$. These are used in the attention operation as follows:

\begin{equation}
\mathcal{A}^{(i)} = \rho\left(\frac{\mathbf{Q}_i\mathbf{K}_i^T}{\sqrt{d_k}} \right) \mathbf{V}_i,
\label{eq:self-att}
\end{equation}

\paragraph{Molecule Self-Attention.}

Using a naive Transformer architecture would require encoding of chemical molecules as sentences. Instead, inspired by \citet{Battaglia2018}, we interpret the self-attention as a soft adjacency matrix between the elements of the input sequence. Following this line of thought, it is natural to augment the self-attention using information about the actual structure of the model. This allows us to avoid using linearized (textual) representation of molecule as input~\cite{jastrzebski2016}, which we expect to be a better inductive bias for the model.

More concretely, we propose the Molecule Self-Attention layer, which we describe in Equation~\ref{eq:mmselfatt}. We augment the self-attention matrix as follows: let $\mathbf{A} \in \{0,1\}^{N_{\text{atoms}} \times N_{\text{atoms}}}$ denote the graph adjacency matrix, and $\mathbf{D} \in \mathbb{R}^{N_{\text{atoms}} \times N_{\text{atoms}}}$. denote the inter-atomic distances. Let $\lambda_{a}$,  $\lambda_{d}$, and $\lambda_{g}$ denote scalars weighting the self-attention, distance, and adjacency matrices. We modify Equation~\ref{eq:self-att} as follows:

\begin{equation}
    \mathcal{A}^{(i)} = \left(\lambda_{a} \rho \left(\frac{\mathbf{Q}_i\mathbf{K}_i^T}{\sqrt{d_k}}\right) + \lambda_d g(\mathbf{D}) + \lambda_g \mathbf{A} \right) \mathbf{V}_i,
    \label{eq:mmselfatt}
\end{equation}

see also Figure~\ref{fig:architecture}. We denote $\lambda_{a}$,  $\lambda_{d}$, and $\lambda_{g}$ jointly as $\mathbf{\lambda}$. We use as $g$ either softmax (normalized over the rows), or an element-wise $g(d)=\exp(-d)$. Finally, the distance matrix  $\mathbf{D}$ is computed using RDKit package~\citep{rdkit2016}. 

Note that while we use only the adjacency and the distance matrices, MAT can be easily extended to include other types of information, e.g. forces between the atoms. 

\paragraph{Molecule Attention Transformer.} 

To define the model, we replace all self-attention layers in the original Transformer encoder by our Molecular Self Attention layers. We embed each atom as a $26$ dimensional vector following~\citep{coley2017}, shown in Table~\ref{tab:featurization_atoms}. In the experiments, we treat  $\lambda_{a}$,  $\lambda_{d}$, and $\lambda_{g}$ as hyperparameters and keep them frozen during training. Figure ~\ref{fig:architecture} illustrates the model.

\paragraph{Pretraining.} 

We experiment with one of the two \emph{node-level} pretraining tasks proposed in \citet{Hu2019}, which involves predicting the masked input nodes. Consistently with \citet{Hu2019}, we found it stabilizes learning (see Figure \ref{fig:training_stability}) and reduces the need for an extensive hyperparameter search (see Table \ref{tab:results_pretrained}). Given that MAT already achieves good performance using this simple pretraining task, we leave for future work exploring the other tasks proposed in \citet{Hu2019}.

\paragraph{Other details.}  

Inspired by \citet{li2017learning,clark2019does}, we add an artificial \emph{dummy node} to the molecule. The dummy node is not connected by an edge to any other atom and the distance to any of them is set to $10^6$. 
Our motivation is to allow the model to skip searching for a molecular pattern if none is to find by putting higher attention on that distant node, which is similar to how BERT uses the separation token~\citep{devlin2018pretraining,clark2019does}. We confirm this intuition in Section~\ref{sec:ablation} and Section~\ref{sec:analysis}.

Finally, the distance matrices are calculated from 3D conformers calculated using \textsc{UFFOptimizeMolecule} function from the RDKit package~\citep{rdkit2016}, and the default parameters (\textsc{maxIters}=$200$, \textsc{vdwThresh}=$10.0$, \textsc{confId}=$-1$, \textsc{ignoreInterfragInteractions}=True). For each compound we use one pre-computed conformation. We experimented with sampling more conformations for each compound, but did not observe a consistent boost in performance, however it is possible that using more sophisticated algorithms for compound 3D structure minimization could improve the results. We leave this for future work.

{\renewcommand{\arraystretch}{1.3}
\begin{table}[h]
    \centering
    \caption{Featurization used to embed atoms in MAT.}
    \vskip 0.15in
    \small
    \begin{small}
    \begin{sc}
    \begin{tabular}{cc}
        \toprule
         Indices & Description \\
        \midrule
        $0 - 11$ & \begin{tabular}{l}\shortstack{Atomic identity as a one-hot vector of\\  B, N, C, O, F, P, S, Cl, Br, I, Dummy, other\vspace{1mm}}\end{tabular}  \\
        $12 - 17$ & \begin{tabular}{l}\shortstack{Number of heavy neighbors as one-hot \\ vector of 0, 1, 2, 3, 4, 5\vspace{1mm}}\end{tabular}    \\
        $18 - 22$ &  \begin{tabular}{l}\shortstack{Number of hydrogen atoms as \\ one-hot vector of 0, 1, 2, 3, 4 \vspace{1mm}}\end{tabular}   \\
        $23$ & Formal charge \\
        $24$ & Is in a ring  \\
        $25$ & Is aromatic  \\
        \bottomrule
    \end{tabular}
    \end{sc}
    \end{small}
    \label{tab:featurization_atoms}
\end{table}
}

\section{Experiments}

We begin by comparing MAT to other popular models in the literature on a wide range of tasks. We find that with simple pretraining MAT outperforms other methods, while using a small budget for hyperparameter tuning.

In the rest of this section we try to develop understanding of what makes MAT work well. In particular, we find that individual heads in the multi-headed self-attention layers learn chemically interpretable functions.

\subsection{Experimental settings}

Comparing different models for molecule property prediction is challenging. Despite considerable efforts, the community still lacks a standardized way to compare different models. In our work, we use a similar setting to MoleculeNet~\citep{Wu2018}.

\paragraph{Evaluation.} 

Following recommendations of \citet{Wu2018} and the experimental setup of \citet{podlewska2018metstabon}, we use random split for FreeSolv, ESOL, and MetStab. For all the other datasets we use scaffold split, which assigns compounds that share the same molecular scaffolding to different subsets of the data~\cite{bemis1996properties}. In regression tasks, the property value was standardized. Test performance is based on the model which gave best results in the validation setting. Each training was repeated $6$ times, on different train/validation/test splits. All the other experimental details are reported in the Supplement. 

\paragraph{Datasets.} 

We run experiments on a wide range of datasets that represent typical tasks encountered in molecule modeling. Below, we include a short description of these tasks, and a more detailed description is moved to App.~\ref{app:datasets_details}.

\begin{itemize}
    \item \textbf{FreeSolv, ESOL.} Regression tasks used in \citet{Wu2018} for predicting water solubility in terms of the hydration free energy (FreeSolv) and log solubility in mols per litre (ESOL). The datasets have $642$ and $1128$ molecules, respectively.
    \item \textbf{Blood-brain barrier permeability (BBBP).} Binary classification task used in \citet{Wu2018} for predicting the ability of a molecule to penetrate the blood-brain barrier. The dataset has $2039$ molecules.
    \item \textbf{Estrogen Alpha, Estrogen Beta.} The tasks are to predict whether a compound is active towards a given target (Estrogen-$\alpha$, Estrogen-$\beta$) based on experimental data from the ChEMBL database~\citep{chembl2011}. The datasets have $2398$, and $1961$ molecules, respectively.
    \item \textbf{MetStab\textsubscript{high}, MetStab\textsubscript{low}.} Binary classification tasks based on data from \citet{podlewska2018metstabon} to predict whether a compound has high (over 2.32~h half-time) or low (lower than 0.6~h half-time) metabolic stability. Both datasets contain the same $2127$ molecules.

\end{itemize}

\subsection{Molecule Attention Transformer}
\label{sec:comp_models}

\begin{figure*}[t!] 
    \centering
        \begin{subfigure}[b]{0.45\textwidth}
        \centering
        \includegraphics[width=\textwidth]{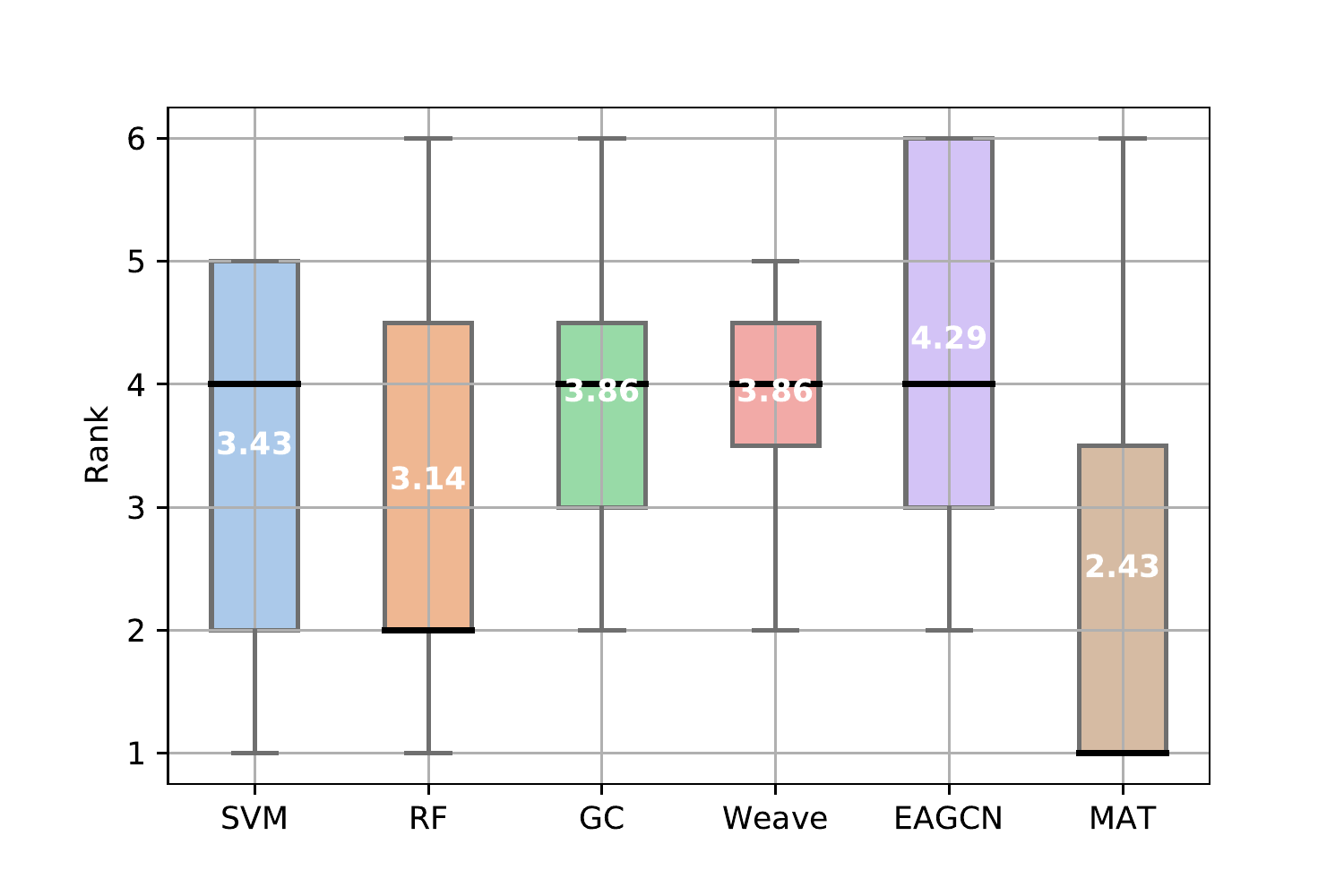}
        \caption{Hyperparameter search budget of 500 combinations.}
        \label{fig:rankplot_500}
    \end{subfigure}%
    \begin{subfigure}[b]{0.45\textwidth}
        \centering
        \includegraphics[width=\textwidth]{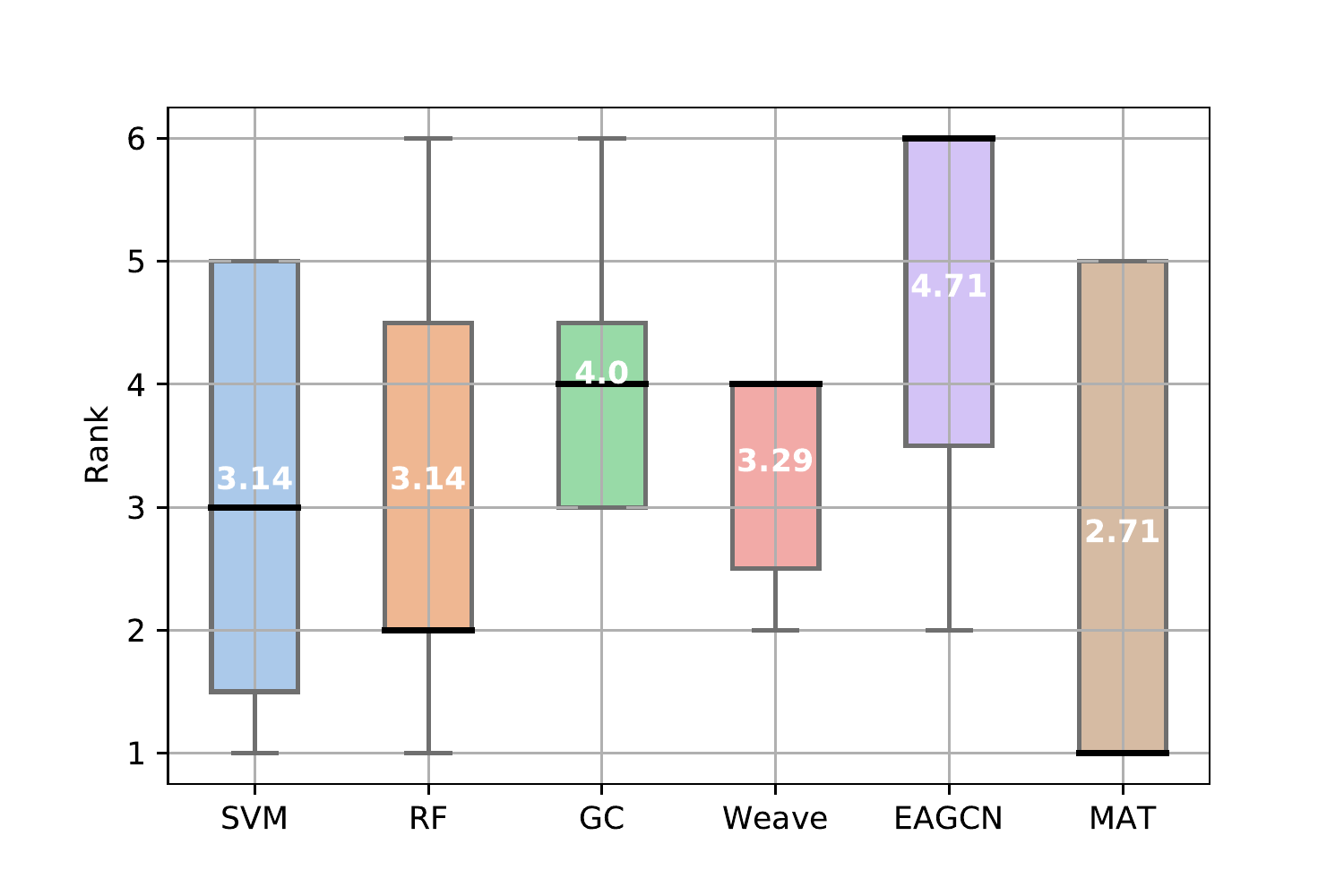}
        \caption{Hyperparameter search budget of $150$ combinations.}
        \label{fig:rankplot_150}
    \end{subfigure}
    \caption{The average rank across the 7 datasets in the benchmark. For each model we test $500$ (left) or $150$ (right) hyperparameter combinations. We split the data using random or scaffold split (according to the dataset description) 6 times into train/validation/test folds and use the mean metrics across the test folds to obtain the ranklists of models. Interestingly, \emph{shallow} models (RF and SVM) outperform graph models (GCN, EAGCN and Weave). }
    \label{fig:main_results}
\end{figure*}

\begin{table*}[h!]
    \centering
    \caption{Test performances in the benchmark. For each model we test $500$ (top) and $150$ (bottom) hyperparameter combinations. On ESOL and FreeSolv we report RMSE (lower is better). The other tasks are evaluated using ROC AUC (higher is better). Experiments are repeated $6$ times. }
    \vskip 0.15in
    \small
    \begin{subtable}{1.0\textwidth}
    \centering
    \caption{Hyperparameter search budget of $500$ combinations.}
    \begin{small}
    \begin{sc}
    \begin{tabular}{lccccccc}
    \toprule
    {} &           BBBP &           ESOL &       FreeSolv & Estrogen-$\alpha$ &  Estrogen-$\beta$ &    MetStab\textsubscript{low} &   MetStab\textsubscript{high}  \\
    \midrule
    SVM    &    .707 $\pm$ .000 &  .478 $\pm$ .054 &  .461 $\pm$ .077 &    .973 $\pm$ .000 &    .778 $\pm$ .000 &   $\mathbf{.893 \pm .030}$ &   $\mathbf{.890 \pm .029}$ \\
    RF     &  .725 $\pm$ .006 &  .534 $\pm$ .073 &  .523 $\pm$ .097 &  .977 $\pm$ .001 &  $\mathbf{.797 \pm .007}$ &  .885 $\pm$ .029 &   .888 $\pm$ .030 \\
    GCN     &   .712 $\pm$ .010 &  .357 $\pm$ .032 &  .271 $\pm$ .048 &  .975 $\pm$ .003 &  .730 $\pm$ .006 &  .881 $\pm$ .031 &  .875 $\pm$ .036 \\
    Weave  &  .701 $\pm$ .016 &  .311 $\pm$ .023 &  .311 $\pm$ .072 &   .974 $\pm$ .003 &  .769 $\pm$ .023 &  .863 $\pm$ .028 &  .882 $\pm$ .043 \\
    EAGCN &   .680 $\pm$ .014 &  .316 $\pm$ .024 &  .345 $\pm$ .051 &  .961 $\pm$ .011 &  .781 $\pm$ .012 &  .883 $\pm$ .024 &  .868 $\pm$ .034 \\
    MAT (ours)    &   $\mathbf{.728 \pm .008}$ &    $\mathbf{.285 \pm .022}$ &    $\mathbf{.263 \pm .046}$ &  $\mathbf{.979 \pm .003}$ &  .765 $\pm$ .007 &    .862 $\pm$ .038 &    .888 $\pm$ .027 \\
    \bottomrule
    \end{tabular}
    \end{sc}
    \end{small}
    \end{subtable}%
    \vspace{.3cm}
    \begin{subtable}{1.0\textwidth}
    \centering
    \caption{Hyperparameter search budget of $150$ combinations.}
    \begin{small}
    \begin{sc}
    \begin{tabular}{lccccccc}
    \toprule
    {} &           BBBP &           ESOL &       FreeSolv & Estrogen-$\alpha$ &  Estrogen-$\beta$ &    MetStab\textsubscript{low} &   MetStab\textsubscript{high} \\
    \midrule
    SVM   &    .723 $\pm$ .000 &  .479 $\pm$ .055 &  .461 $\pm$ .077 &    .973 $\pm$ .000 &    .772 $\pm$ .000 &   $\mathbf{.893 \pm .030}$ &   $\mathbf{.890 \pm .029}$ \\
    RF    &  .721 $\pm$ .003 &  .534 $\pm$ .073 &  .524 $\pm$ .098 &  .977 $\pm$ .001 &  $\mathbf{.791 \pm .012}$ &  .892 $\pm$ .026 &   .888 $\pm$ .030 \\
    GCN    &  .695 $\pm$ .013 &  .369 $\pm$ .032 &  .299 $\pm$ .068 &  .975 $\pm$ .003 &   .730 $\pm$ .006 &  .884 $\pm$ .033 &  .875 $\pm$ .036 \\
    Weave &  .702 $\pm$ .009 &  .298 $\pm$ .025 &  .298 $\pm$ .049 &  .974 $\pm$ .003 &  .769 $\pm$ .023 &  .863 $\pm$ .028 &  .885 $\pm$ .042 \\
    EAGCN &   .680 $\pm$ .014 &  .322 $\pm$ .052 &  .337 $\pm$ .042 &  .961 $\pm$ .011 &  .781 $\pm$ .012 &  .859 $\pm$ .024 &  .844 $\pm$ .037 \\
    MAT (ours)   &  $\mathbf{.727 \pm .006}$ &   $\mathbf{.290 \pm .019}$ &  $\mathbf{.289 \pm .047}$ &  $\mathbf{.979 \pm .003}$ &  .765 $\pm$ .007 &  .861 $\pm$ .029 &  .844 $\pm$ .052 \\
    \bottomrule
    \end{tabular}
    \end{sc}
    \end{small}
    \end{subtable}
    \label{tab:results}
\end{table*}

\begin{figure}[h]
    \centering
    \includegraphics[width=0.45\textwidth]{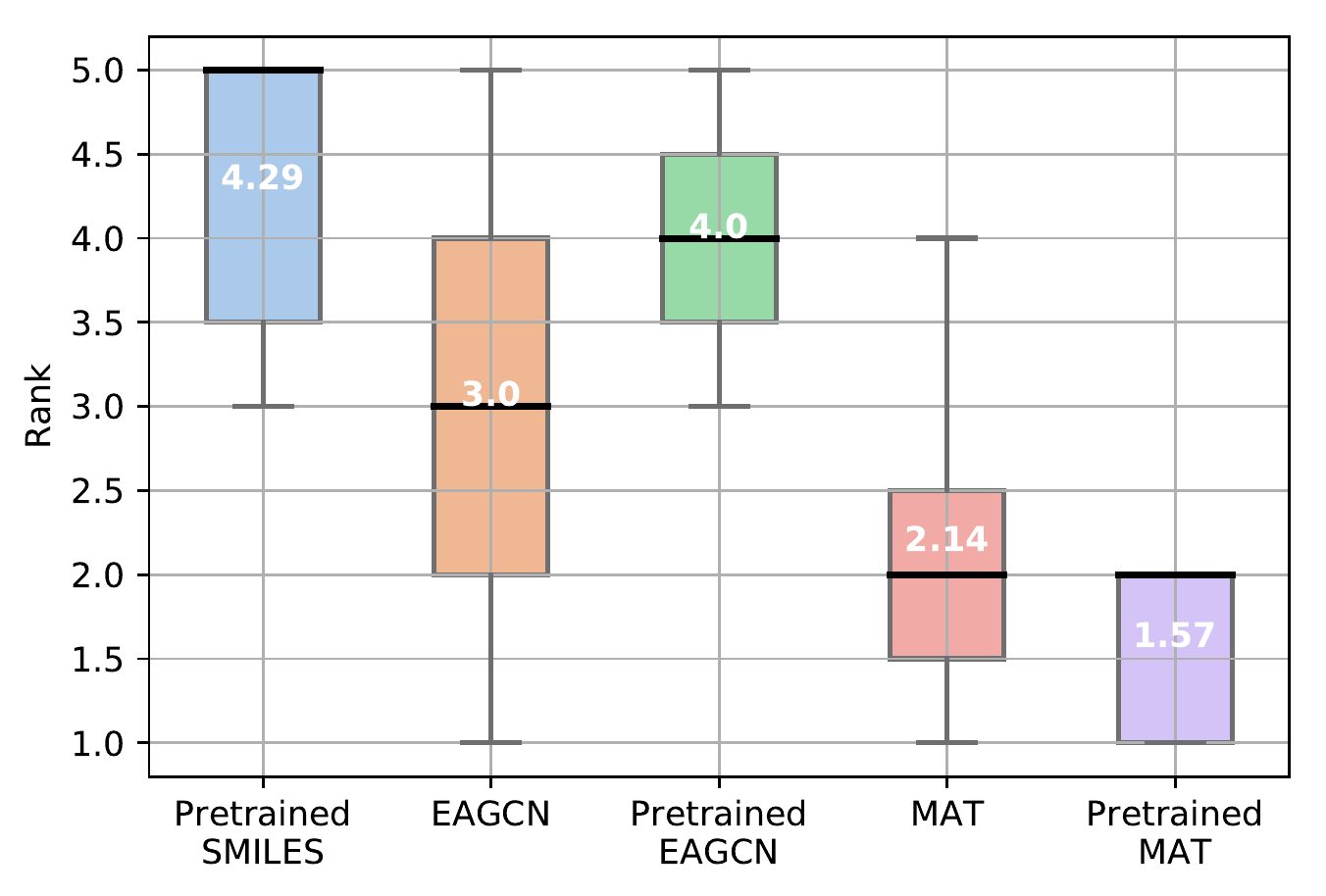}
    \caption{The average ranks across the 7 datasets in the benchmark. Pretrained MAT outperforms the other methods, despite a drastically smaller number of tested hyperparameters ($7$) compared to MAT and EAGCN ($500$).  }
    \label{fig:results_500_pretrained}
\end{figure}

\paragraph{Models.} Similarly to \citet{Wu2018}, we test a comprehensive set of baselines that span both shallow and deep models. We compare MAT to the following baselines: GCN~\cite{Duvenaud2015}, Random Forest (RF) and Support Vector Machine with RBF kernel (SVM). We also test the following recently proposed models: Edge Attention-based Multi-Relational Graph Convolutional Network (EAGCN)~\citep{shang2018}, and \mbox{Weave}~\citep{kearnes2016}.

\paragraph{Hyperparameter tuning.} For each method we extensively tune their hyperparameters using random search~\citep{bergstra2012}. To ensure fair comparison, each model is given the same budget for hyperparameter search. We run two sets of experiments with budget of $150$ and $500$ evaluations. We include hyperparameter ranges in App.~\ref{app:exp_details}.

\paragraph{Results.} 

We evaluate models by their average rank according to the test set performance on the $7$ datasets. Figure~\ref{fig:main_results} reports ranks of all methods for the two considered hyperparameter budgets ($150$ and $500$). Additionally, we report in Table~\ref{tab:results} detailed scores on all datasets. We make three main observations.

First, graph neural networks (GCN, Weave, EAGCN) on average do not outperform the other models. The best graph model achieves average rank $3.28$ compared to $3.14$ by RF. On the whole, performance of the deep models improves with larger hyperparameter search budget. This further corroborates the original motivation of our study. Indeed, using common deep learning methods for molecule property prediction is challenging in practice. It requires a large computational budget, and might still result in poor performance.

Second, MAT outperforms the other tested methods in terms of the average rank. MAT achieves average rank of $2.71$ and $2.42$ for $150$ and $500$ budgets, compared to $3.14$ of RF, which is the second best performing model. This shows that architecture MAT is flexible enough and has the correct inductive bias to perform well on a wide range of tasks. 

Examining performance of MAT across individual datasets, we observe that RF and SVM perform better on \mbox{Estrogen-$\beta$}, MetStab\textsubscript{low}, and MetStab\textsubscript{high}. Both RF and SVM use extended-connectivity fingerprint~\citep{rogers2010extended} as input representation, which encodes substructures in the molecule as features. Metabolic stability of a compound depends on existence of particular moieties, which are recognized by enzymes. Therefore a simple structure-based fingerprints perform well in such a setting. \citet{wang2019,Mayr2018} show that using fingerprint as input representation improves performance of deep networks on related datasets. These two observations suggest that MAT could benefit from using fingerprints. Instead, we avoid using handcrafted representations, and investigate pretraining as an alternative in the next section. Though fingerprint-based models show excellent performance in all presented tasks, there are datasets on which they fail to match the performance of graph approaches. We observed this also on an energy prediction task (see the extension of our benchmark in App.~\ref{app:more_experiments}).

\subsection{Pretrained Molecule Attention Transformer}
\label{sec:pretraining}

Self-supervised pretraining has revolutionized natural language processing~\citep{devlin2018pretraining} and has improved performance in molecule property prediction~\citep{Hu2019}. We apply here node-level self-supervised pretraining from \citet{Hu2019} to MAT. The task is to predict features of masked out nodes. We refer the reader to App.~\ref{app:pretraining} for more details.

\paragraph{Models.} We compare MAT to the two following baselines. First, we apply the same pretraining to EAGCN, which we will refer to as ``Pretrained EAGCN''. Second, we compare to a concurrent work by \citet{honda2019smiles}. They pretrain a vanilla Transformer by decoding textual representation (SMILES) of molecules. We will refer to their method as ``SMILES Transformer''.

\paragraph{Hyperparameters.} For all methods that use pretraining we reduce the hyperparameter grid to a minimum. We tune only the learning rate in $\{ 1e{-3}, 5e{-4}, 1e{-4}, 5e{-5}, 1e{-5}, 5e{-6}, 1e{-6} \}$. We set the other hyperparameters to reasonable defaults based on results from Section ~\ref{sec:comp_models}. For MAT and EAGCN, we follow \cite{devlin2018pretraining} and use the largest model that still fits the GPU memory. For SMILES Transformer we use pretrained weights provided by \citet{honda2019smiles}.

\paragraph{Results.}

As in previous section, we compare the models based on their average rank on our benchmark. Figure ~\ref{fig:results_500_pretrained} and Table~\ref{tab:results_pretrained} summarize the results.

\begin{figure*}[t!] 
    \centering
        \begin{subfigure}[b]{0.49\textwidth}
        \centering
        \includegraphics[width=\textwidth]{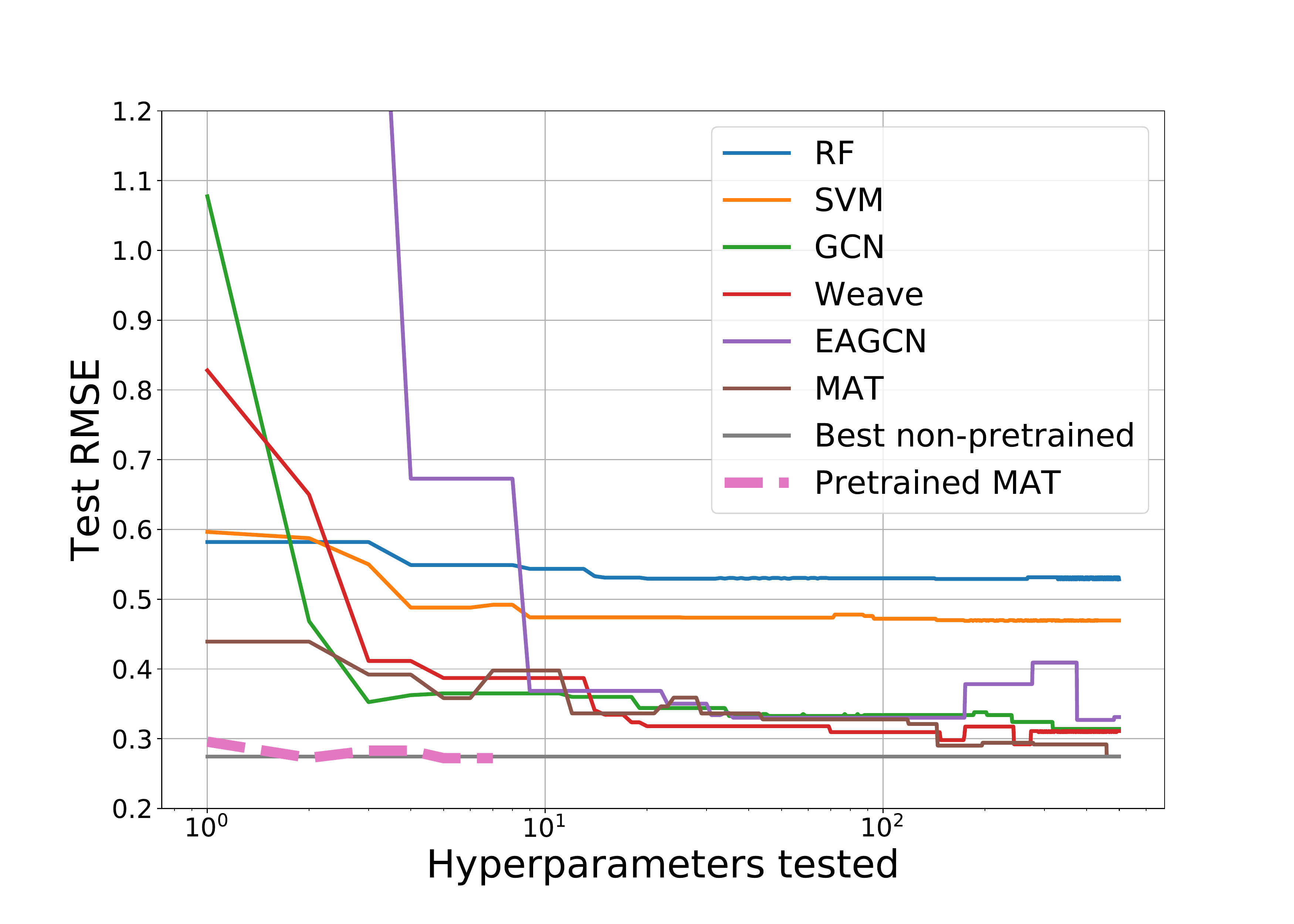}
        \caption{Regression tasks.}
        \label{fig:hps_regression}
    \end{subfigure}%
    \begin{subfigure}[b]{0.49\textwidth}
        \centering
        \includegraphics[width=\textwidth]{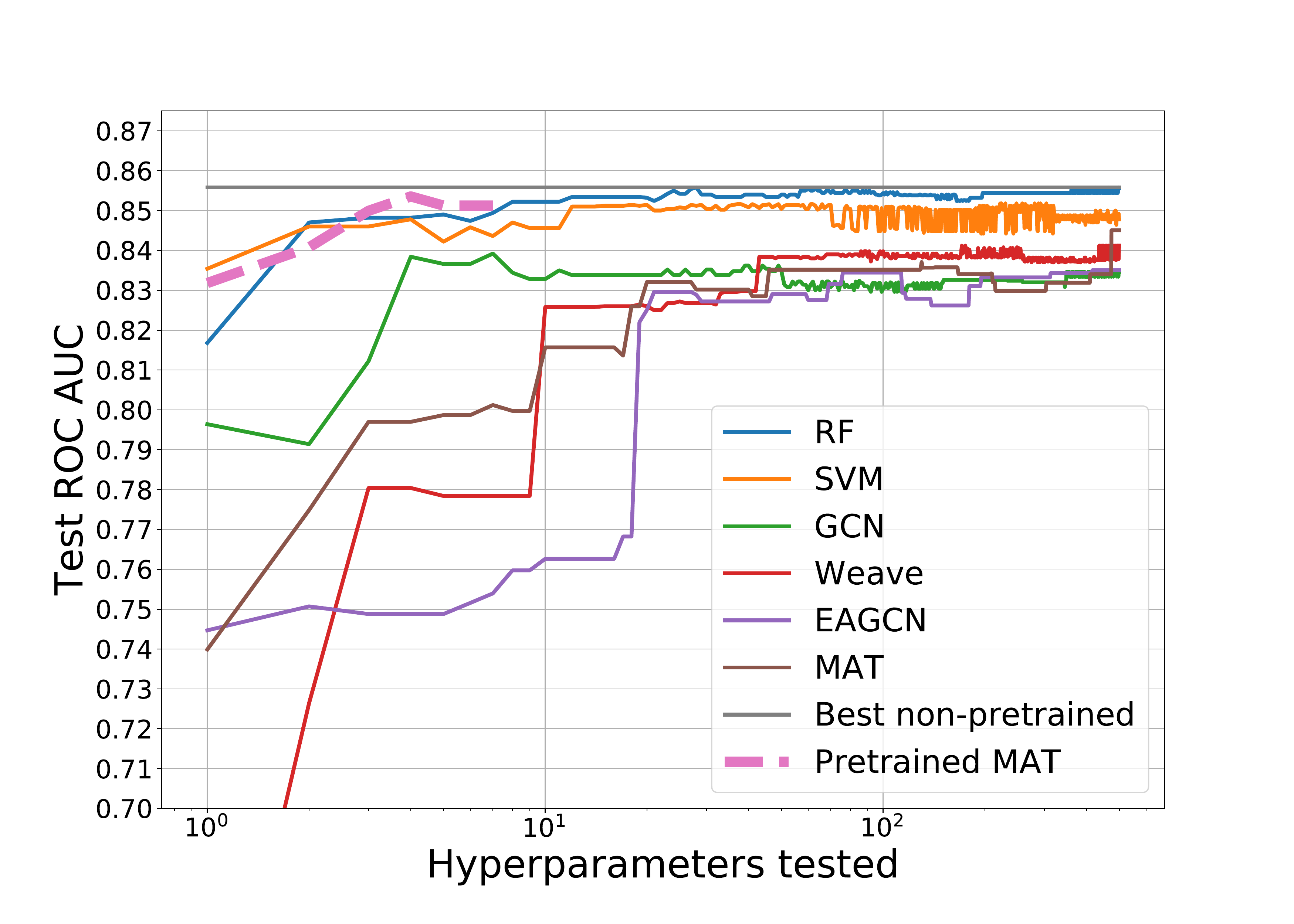}
        \caption{Classification tasks.}
        \label{fig:hps_classification}
    \end{subfigure}
    \caption{Test performance of all models as a function of the number of tested hyperparameter combinations (on a logarithmic scale). Figures show the aggregated mean RMSE for regression tasks (left) and the aggregated mean ROC AUC for classification tasks (right). Pretrained MAT requires tuning an order of magnitude less hyperparameters, and performs competitively on both sets of tasks.}
    \label{fig:hps_plots}
\end{figure*}

We observe that Pretrained MAT achieves average rank of $1.57$ and outperforms MAT (average rank of $2.14$). Importantly, for Pretrained MAT we only tuned the learning rate by evaluating $7$ different values. This is in stark contrast to the $500$ hyperparameter combinations tested for MAT and EAGCN. To visualize this, in Figure~\ref{fig:hps_plots} we plot the average test performance of all models as a function of the number of tested hyperparameter combinations. We also note that Pretrained MAT is more competitive on the three datasets mentioned in the previous section.

We also find that Pretrained MAT outperforms the other two pretrained methods. Pretraining degrades the performance of EAGCN (average rank of $4.0$), and SMILES Transformer achieves the worst average rank (average rank of $4.29$). This suggests that both the architecture, and the choice of the pretraining task are important for the overall performance of the model.

\begin{table*}[h]
    \centering
        \caption{Test set performances of methods that use pretraining. Experiments are repeated $6$ times. SMILES refers to SMILES Transformer from \citet{honda2019smiles}.}
        \vskip 0.15in
    \begin{small}
    \begin{sc}
    \begin{tabular}{lccccccc}
    \toprule
    {} &           BBBP &           ESOL &       FreeSolv & Estrogen-$\alpha$ &  Estrogen-$\beta$ &    MetStab\textsubscript{low} &   MetStab\textsubscript{high} \\
    \midrule
    MAT   &  $\mathbf{.737 \pm .009}$ &   $\mathbf{.278 \pm .020}$ &   $\mathbf{.265 \pm .042}$ &    $\mathbf{.998 \pm .000}$ &  $\mathbf{.773 \pm .012}$ &  $\mathbf{.862 \pm .025}$ &  $\mathbf{.884 \pm .030}$ \\
    EAGCN &  .687 $\pm$ .023 &  .323 $\pm$ .031 &  1.244 $\pm$ .341 &  .994 $\pm$ .002 &    .770 $\pm$ .010 &  .861 $\pm$ .029 &  .839 $\pm$ .038 \\
    SMILES &  .717 $\pm$ .008 &  .356 $\pm$ .017 &  .393 $\pm$ .032 &  .953 $\pm$ .002 &    .757 $\pm$ .002 &  .860 $\pm$ .038 &  .881 $\pm$ .036 \\
    \bottomrule
    \end{tabular}
    \end{sc}
    \end{small}
    \label{tab:results_pretrained}
    \vskip -0.1in
\end{table*}

\subsection{Ablation studies}
\label{sec:ablation}

To better understand what contributes to the performance of MAT, we run a series of ablation studies on three representative datasets from our benchmark. We leave understanding how these choices interact with pretraining for future work. 

For experiments in this section we generated additional splits for ESOL, FreeSolv and BBBP datasets (different than in Section \ref{sec:comp_models}). For each configuration we select the best hyperparameters settings using random search under a budget of $100$ evaluations. Experiments are repeated $3$ times.

\paragraph{Dummy node is not so dummy.}

\begin{table}
    \centering
        \caption{Test performance of MAT model variant without the dummy node (- \textsc{Dummy}) compared to performance of the original MAT. }
        \vskip 0.15in
    \begin{small}
    \begin{sc}
    \begin{tabular}{lccc}
    \toprule
    {} &           BBBP &           ESOL &       FreeSolv \\
    \midrule
    MAT & $\mathbf{.723 \pm .008}$ & $\mathbf{.286 \pm .006}$ & .250 $\pm$ .007 \\
    - Dummy & .714 $\pm$ .010 & .317 $\pm$ .014 & $\mathbf{.249 \pm .014}$ \\ 
    \bottomrule
    \end{tabular}
    \end{sc}
    \end{small}
    \label{tab:ablation_dummy_node}
    \vskip -0.1in
\end{table}

MAT uses a \emph{dummy node} that is disconnected from other atoms in the graph~\citep{li2017learning}. Our intuition is that such functionality can be useful to automatically adapt capacity on small datasets. By attending to the dummy node, the model can effectively choose to avoid changing the internal representation in a given layer. To examine this architectural choice, in Table~\ref{tab:ablation_dummy_node} we compare MAT to a variant that does not include the dummy node. Results show that dummy node improves performance of the model.

\paragraph{Knowing molecular graph and distances between atoms improves performance.} Our key architectural innovation is integrating the molecule graph and inter-atomic distances with the self-attention layer in Transformer, as shown in Figure~\ref{fig:architecture}. To probe the importance of each of these sources of information, we removed them individually during training. Results in Table~\ref{tab:ablation_lambdas} suggest that keeping all sources of information results in the most stable performance across the three tasks, which is our primary goal. We also show that MAT can effectively use distance information in a toy task involving 3-dimensional distances between functional groups (see~App.\ref{app:toy}).

\begin{table}
    \centering
        \caption{Test performance of MAT with different sources of information removed (equivalent to setting the corresponding $\lambda$ to zero). }
        \vskip 0.15in
    \begin{small}
    \begin{sc}
    \begin{tabular}{lccc}
    \toprule
    {} &           BBBP &           ESOL &       FreeSolv \\
    \midrule
    MAT & .723 $\pm$ .008 & .286 $\pm$ .006 & $\mathbf{.250 \pm .007}$ \\
    - graph & .716 $\pm$ .009 & .316 $\pm$ .036 & .276 $\pm$ .034 \\
    - distance & $\mathbf{.729 \pm .013}$ & $\mathbf{.281 \pm .001}$ & .281 $\pm$ .013 \\
    - attention & .692 $\pm$ .001 & .306 $\pm$ .026 & .329 $\pm$ .014 \\
    \bottomrule
    \end{tabular}
    \end{sc}
    \end{small}
    \label{tab:ablation_lambdas}
    \vskip -0.1in
\end{table}

\paragraph{Using a more complex featurization does not improve performance.}

Many models for predicting molecule properties use additional edge features~\citep{coley2017, shang2018, gilmer2017}. In Table~\ref{tab:ablation_edge} we show that adding additional edge features does not improve MAT performance. This is certainly possible that a more comprehensive set of edge features or a better method to integrate them would improve performance, which we leave for future work. Procedure of using edge features is described in detail in App. \ref{app:ablation_appendix}.

\begin{table}
    \centering
        \caption{Test performance of MAT using additional edge features (+ \textsc{Edges f.}), compared to vanilla MAT. }
        \vskip 0.15in
    \begin{small}
    \begin{sc}
    \begin{tabular}{lccc}
    \toprule
    {} &           BBBP &           ESOL &       FreeSolv \\
    \midrule
    MAT & $\mathbf{.723 \pm .008}$ & $\mathbf{.286 \pm .006}$ & $\mathbf{.250 \pm .007}$ \\
    + Edges f. & .683 $\pm$ .008 & $.314 \pm .014$ & $.358 \pm .023$ \\
    \bottomrule
    \end{tabular}
    \end{sc}
    \end{small}
    \label{tab:ablation_edge}
    \vskip -0.1in
\end{table}

\subsection{Analysis.}
\label{sec:analysis}

To understand MAT better, we investigate attention weights of the model, and the effect of pretraining on the learning dynamics. 

\paragraph{What is MAT looking at?} In natural language processing, it has been shown that heads in Transformer seem to implement interpretable functions~\citep{htut2019attention,clark2019does}. Similarly, we investigate here the chemical function implemented by self-attention heads in MAT. We show patterns found in the model that was pretrained with the atom masking strategy~\citep{Hu2019}, and then we verify our findings on a set of molecules extracted from the BBBP testing dataset.

\begin{figure}[h!]
  \centering
  \includegraphics[width=.45\textwidth]{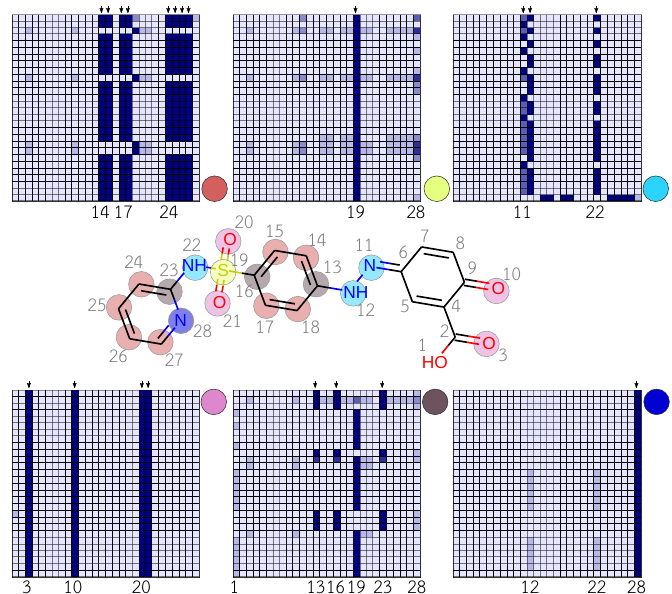}%
  \caption{The heatmaps show selected self-attention weights from the first layer of MAT, on a random molecule from the BBBP dataset (center). The atoms, which these heads focus on, are marked with the same color as the corresponding matrix. The interpretation of the presented patterns is described in the text.}
  \label{fig:heads}
\end{figure}

Based on a manual inspection of attention matrices of MAT, we find two broad patterns: (1) many attention heads are almost fully focused on the dummy node, (2) many attention heads focus only on a few atoms. This seems consistent with observations about Transformer in \citet{clark2019does}. We also notice that initial self-attention layers learn simple and easily interpretable chemical patterns, while subsequent layers capture more complex arrangements of atoms. 
In Figure~\ref{fig:heads} we exemplify attention patterns on a random molecule from the BBBP dataset.

To quantify the above findings, we select six heads from the first layer that fit the second category and seem to implement six patterns: (i) focuses on 2-neighboured aromatic carbons (not substituted); (ii) focuses on sulfurs; (iii) focuses on non-ring nitrogens; (iv) focuses on oxygen in carbonyl groups; (v) focuses on 3-neighboured aromatic atoms (positions of aromatic ring substitutions) and on sulfur for different atoms; (vi) focuses on nitrogens in aromatic rings. We found that on the BBBP testing dataset the atoms corresponding to these definitions (queried with SMARTS expressions) have indeed higher attention weights assigned to them than other atoms. For each head, we calculated attention weights for all atoms in all molecules and compared those matching our hypothesis against the other atoms. Their distributions differ significantly ($p<0.001$ in Kruskal-Wallis test) for all the patterns. The statistics and experimental details are summarized in App.~\ref{app:analysis}.

\begin{figure}[h!]
    \centering
    \begin{subfigure}[b]{0.24\textwidth}
        \centering
        \includegraphics[width=\textwidth]{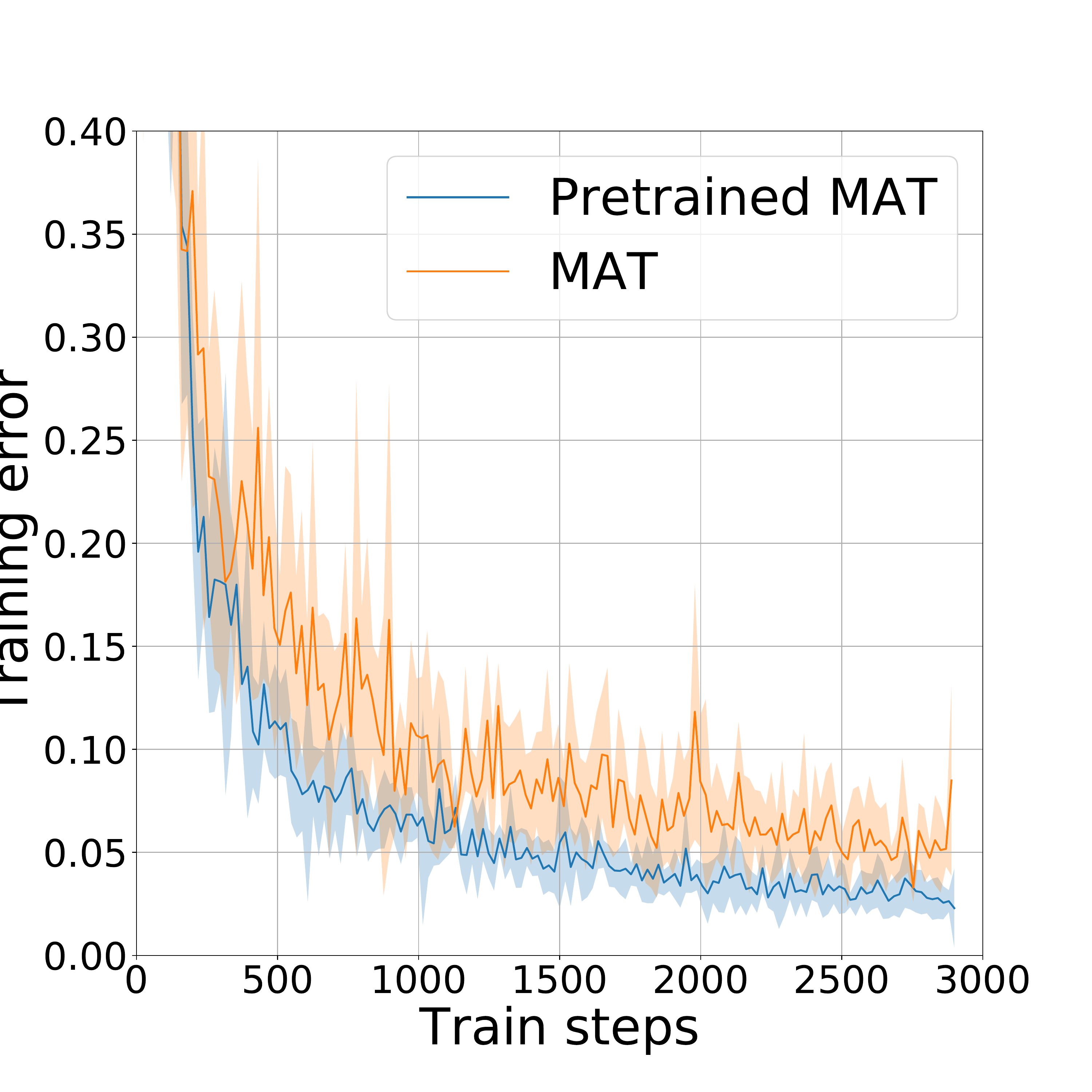}
        \caption{ESOL}
        \label{fig:esol_curve}
    \end{subfigure}%
    \begin{subfigure}[b]{0.24\textwidth}
        \centering
        \includegraphics[width=\textwidth]{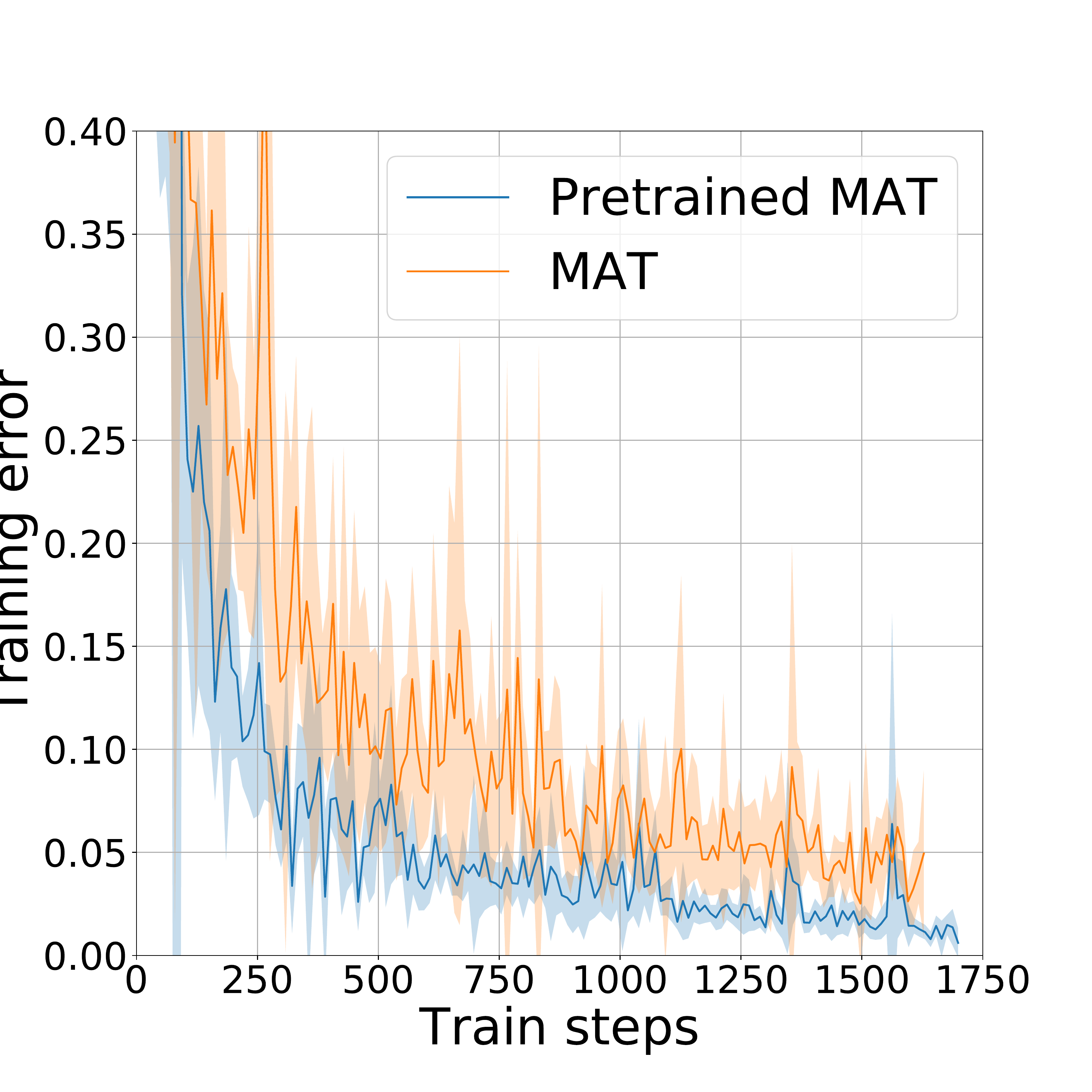}
        \caption{FreeSolv}
        \label{fig:freesolv_curve}
    \end{subfigure}
    \caption{Training of MAT with (blue) and without (orange) pretraining, on ESOL (left) and FreeSolv (right). Pretraining stabilizes training (smaller variance of the training error) and improves convergence speed.}
    \label{fig:training_stability}
\end{figure}

\paragraph{Effect of pretraining.} \citet{Wu2018} observed that using pretraining stabilizes and speeds up training of graph convolutional models. We observe a similar effect in our case. Figure~\ref{fig:training_stability} reports training error of MAT and Pretrained MAT on the ESOL (left), and the FreeSolv (right) datasets. We use the learning rate that achieved the best generalization on each dataset in Sec.~\ref{sec:pretraining}. The experiments are repeated $6$ times. On both datasets, Pretrained MAT converges faster and has a lower variance of training error across repetitions. Mean standard deviation of training error for Pretrained MAT (MAT) is $0.027$ ($0.057$) and $0.040$ ($0.076$) for ESOL and FreeSolv, respectively. 

\section{Conclusions.}

In this work we propose Molecule Attention Transformer as a versatile architecture for molecular property prediction. In contrast to other tested models, MAT performs well across a wide range of molecule property prediction tasks. Moreover, inclusion of self-supervised pretraining further improves its performance, and drastically reduces the need for tuning of hyperparameters.

We hope that our work will widen adoption of deep learning in applications involving molecular property prediction, as well as inspire new modeling approaches. One particularly promising avenue for future work is exploring better pretraining tasks for MAT.

\bibliography{main}
\bibliographystyle{preprint}

\clearpage
\appendix

\section{Dataset details.}
\label{app:datasets_details}

We include below a more detailed description of the datasets used in our benchmark.

\begin{itemize}
    \item \textbf{FreeSolv, ESOL.} Regression tasks. Popular tasks for predicting water solubility in terms of the hydration free energy (FreeSolv) and logS (ESOL). Solubility of molecules is an important property that influences the bioavailability of drugs.
    \item \textbf{Blood-brain barrier permeability (BBBP).} Binary classification task. The blood-brain barrier (BBB) separates the central nervous system from the bloodstream. Predicting BBB penetration is especially relevant in drug design when the goal for the molecule is either to reach the central nervous system or the contrary -- not to affect the brain. 
    \item \textbf{MetStab\textsubscript{high}, MetStab\textsubscript{low}.} Binary classification tasks. The metabolic stability of a compound is a measure of the half-life time of the compound within an organism. The compounds for this task were taken from~\citep{podlewska2018metstabon}, where compounds were divided into three sets: high, medium, and low stability. In this paper we concatenated these sets in order to build two classification tasks: MetStab\textsubscript{high} (discriminating high against others) and MetStab\textsubscript{low} (discriminating low against others).
    \item \textbf{Estrogen Alpha, Estrogen Beta.} Binary classification tasks. Often in drug discovery, it is important that a molecule is not potent towards a given target. Modulating of the estrogen receptors changes the genomic expression throughout the body, which in turn may lead to the development of cancer. For these tasks, the compounds with known activities towards the receptors were extracted from ChEMBL~\citep{chembl2011} database and divided into active and inactive sets based on their reported inhibition constant (Ki), being $<100$ nM and $>1000$ nM, respectively.
\end{itemize}

\section{Other experimental details}
\label{app:exp_details}

In this section we include details for hyperparameters and training settings used in Section~\ref{sec:comp_models}.

\paragraph{Molecule Atention Trainsformer.}

Table~\ref{app:hp_MAT} shows hyperparameter ranges used in experiments for MAT. A short description of these hyperparameters is listed below:

\begin{itemize}
    \item \textsc{model dim} -- size of embedded atom features,
    \item \textsc{layers number} -- number of encoder module repeats ($\mathbf{N}$ in Figure \ref{fig:architecture}),
    \item \textsc{attention heads number} -- number of molecule self-attention heads,
    \item \textsc{PFFs number} -- number of dense layers in the position-wise feed forward block ($\mathbf{K}$ in Figure \ref{fig:architecture}),
    \item \textsc{$\lambda_{att}$} -- self-attention weight $\lambda_{att}$,
    \item \textsc{$\lambda_{dist}$} -- distance weight $\lambda_d$,
    \item \textsc{distance matrix kernel} -- function $g$ used to transform the distance matrix $\mathbf{D}$,
    \item \textsc{model dropout} -- dropout applied after the embedding layer, position-wise feed forward layers, and residual layers (before sum operation),
    \item \textsc{weight decay} -- optimizer weight decay,
    \item \textsc{learning rate} -- (see Equation \ref{eq:optimizer})
    \item \textsc{epochs number} -- number of epochs for which the model is trained
    \item \textsc{batch size} -- batch size used during the training of the model
    \item \textsc{warmup factor} -- fraction of epochs after which we end with increasing the learning rate linearly and begin with decreasing it proportionally to the inverse square root of the step number. (see Equation \ref{eq:optimizer})
\end{itemize}

\begin{table}[h]
\centering
\caption{Molecule Attention Transformer hyperparameters ranges}
\vskip 0.15in
\begin{small}
\begin{sc}
    \begin{tabular}{ll}
    \toprule
    {} &                                         parameters \\
    \midrule
    batch size                      &                               8, 16, 32, 64, 128 \\
    learning rate                   &                    .01, .005, .001, .0005, .0001 \\
    epochs                          &                                          30, 100 \\
    model dim                       &                      32, 64, 128, 256, 512, 1024 \\
    layers number                   &                                    1, 2, 4, 6, 8 \\
    attention heads number          &                                   1, 2, 4, 8, 16 \\
    PFFs number                     &                                                1 \\
    $\lambda_{att}$                 &         0, .1, .2, .3, .4, .5, .6, .7, .8, .9, 1 \\
    $\lambda_{distance}$            &         0, .1, .2, .3, .4, .5, .6, .7, .8, .9, 1 \\
    distance matrix kernel          &                                 'softmax', 'exp' \\
    model dropout                   &                                       .0, .1, .2 \\
    weight decay                    &                     .0, .00001, .0001, .001, .01 \\
    warmup factor                   &                           .0, .1, .2, .3, .4, .5 \\
    \bottomrule
    \end{tabular}
\end{sc}
\end{small}
\label{app:hp_MAT}
\vskip -0.1in
\end{table}

As suggested in~\cite{vaswani2017}, for optimization of MAT we used Adam optimizer~\cite{kingma2014adam}, with learning rate scheduler given by the following formula:

\begin{equation}
\label{eq:optimizer}
\begin{aligned}
    Step_{LR} = optimizer\ factor \cdot model\ dim^{-0.5} \cdot \\
    \cdot \min \left( step\ num^{-0.5}, step\ num \cdot warmup\ steps^{-0.5} \right).
\end{aligned}
\end{equation}

Where \textit{optimizer factor} is given by $100 \ \cdot$ \textsc{learning rate} and \textit{warmup steps} is given by \textsc{warmup factor} $\cdot$ \textit{total train steps number}.

After $N$ layers embedding of the molecule is calculated by taking the mean of returned by the network vector representations of all atoms (\textit{Global pooling} in Figure \ref{fig:architecture}). Then it is passed to the single linear layer, which returns the prediction.

\paragraph{SVM, RF, GCN, Weave.}
In our experiments, we used DeepChem~\citep{Ramsundar-et-al-2019} implementation of baseline algorithms (SVM, RF, GCN, Weave). We used the same hyperparameters for tuning as were used in DeepChem, having regard to their proposed default values (we list them in Tables \ref{app:hp_svm} - \ref{app:hp_Wave}).

RF and SVM work on the vector representation of molecule given by the Extended-connectivity fingerprints~\citep{rogers2010extended}. ECFP vectors were calculated using class \textsc{CircularFingerprint} from the DeepChem package, with default parameters (\textsc{radius}=2, \textsc{size=2048}).

\begin{table}[h]
\centering
\caption{SVM hyperparameter ranges}
\vskip 0.15in
\begin{small}
\begin{sc}
    \begin{tabular}{ll}
        \toprule
        {} &                                         parameters \\
        \midrule
        C     &  \begin{tabular}{l}\shortstack{.25, .4375, .625, .8125, 1., 1.1875, \\ 1.375, 1.5625, 1.75, 1.9375, 2.125, \\ 2.3125, 2.5, 2.6875, 2.875, 3.0625, \\ 3.25, 3.4375, 3.625, 3.8125, 4. \vspace{1mm}}\end{tabular}  \\
        gamma &  \begin{tabular}{l}\shortstack{.0125, .021875, .03125, .040625, \\ .05, .059375, .06875, .078125, .0875, \\ .096875, .10625, .115625, \\ .125, .134375, .14375,  .153125, \\ .1625, .171875, .18125, .190625, .2 \vspace{1mm}}\end{tabular} \\
        \bottomrule
    \end{tabular}
\end{sc}
\end{small}
\vskip -0.1in
\label{app:hp_svm}
\end{table}

\begin{table}[h]
\centering
\caption{RF hyperparameter ranges}
\vskip 0.15in
\begin{small}
\begin{sc}
    \begin{tabular}{ll}
    \toprule
        {} &                                         parameters \\
        \midrule
        n estimators & \begin{tabular}{l}\shortstack{125, 218, 312, 406, 500, 593, \\ 687, 781, 875, 968, 1062, 1156, \\ 1250, 1343, 1437, 1531, 1625, \\ 1718, 1812, 1906, 2000 }\end{tabular}\\
        \bottomrule
    \end{tabular}
\end{sc}
\end{small}
\vskip -0.1in
    \label{app:hp_rf}
\end{table}

\begin{table}[h]
\centering
\caption{GCN hyperparameter ranges}
\vskip 0.15in
\begin{small}
\begin{sc}
    \begin{tabular}{ll}
        \toprule
        {} &              parameters \\
        \midrule
        batch size              &          64, 128, 256 \\
        learning rate           &  0.002, 0.001, 0.0005 \\
        n filters               &     64, 128, 192, 256 \\
        n fully connected nodes &         128, 256, 512 \\
        \bottomrule
    \end{tabular}
\end{sc}
\end{small}
\vskip -0.1in
\label{app:hp_gc}
\end{table}

\begin{table}[h]
\centering
\caption{Weave hyperparameter ranges}
\vskip 0.15in
\begin{small}
\begin{sc}
    \begin{tabular}{ll}
        \toprule
        {} &                       parameters \\
        \midrule
        batch size    &                16, 32, 64, 128 \\
        nb epoch      &            20, 40, 60, 80, 100 \\
        learning rate &  0.002, 0.001, 0.00075, 0.0005 \\
        n graph feat  &           32, 64, 96, 128, 256 \\
        n pair feat   &                             14 \\
        \bottomrule
    \end{tabular}
\end{sc}
\end{small}
\vskip -0.1in
\label{app:hp_Wave}
\end{table}

\paragraph{EAGCN}

Table~\ref{app:hp_EAGCN} shows hyperparameter ranges used in experiments for EAGCN. For EAGCN with \textit{weighted} structure number of convolutional features $n\_sgc = n\_sgc\_1 + n\_sgc\_2 + n\_sgc\_3 + n\_sgc\_4 + n\_sgc\_5$.

\begin{table}[h]
\centering
\caption{EAGCN hyperparameter ranges}
\vskip 0.15in
\begin{small}
\begin{sc}
    \begin{tabular}{ll}
        \toprule
        {} &                            parameters \\
        \midrule
        batch size      &           16, 32, 64, 128, 256, 512 \\
        EAGCN structure &               'concate', 'weighted' \\
        num epochs      &                             30, 100 \\
        learning rate   &       .01, .005, .001, .0005, .0001 \\
        dropout         &                          .0, .1, .3 \\
        weight decay    &                .0, .001, .01, .0001 \\
        n conv layers   &                          1, 2, 4, 6 \\
        n dense layers  &                          1, 2, 3, 4 \\
        n sgc 1         &                              30, 60 \\
        n sgc 2         &                   5, 10, 15, 20, 30 \\
        n sgc 3         &                   5, 10, 15, 20, 30 \\
        n sgc 4         &                   5, 10, 15, 20, 30 \\
        n sgc 5         &                   5, 10, 15, 20, 30 \\
        dense dim       &                     16, 32, 64, 128 \\
        \bottomrule
    \end{tabular}
\end{sc}
\end{small}
\vskip -0.1in
\label{app:hp_EAGCN}
\end{table}

\section{Additional results for Sec.~\ref{sec:comp_models}}
\label{app:more_experiments}

\paragraph{Predicting internal energy} We run an additional experiment on a regression task related to quantum mechanics. From the Alchemy dataset~\citep{chen2019alchemy}, which is a dataset of 12 quantum properties calculated for 200K molecules, we have chosen internal energy at 298.15 K to further test the performance of our model. We hypothesize that our molecule self-attention should perform particularly well in tasks involving atom level interactions such as energy prediction.

Table~\ref{tab:alchemy} presents mean absolute errors of three methods: one classical method (RF), one graph method (GCN), and our pretrained MAT. We use original train/valid/test splits of the dataset. For RF and GCN we run a random search with budget of 500 hyperparameter sets. For pretrained MAT, we tune only the learning rate, that is selected from $\{ 1e{-3}, 5e{-4}, 1e{-4}, 5e{-5}, 1e{-5}, 5e{-6}, 1e{-6} \}$.

MAT achieves a slightly lower error than GCN. As can be expected, both graph methods can learn internal energy function correctly because of the locality preserved in the graph structure. The classical method based on fingerprints gives MAE that is almost two orders of magnitude higher than MAE of the other methods in the comparison.

\begin{table}[h]
    \centering
        \caption{Test results for internal energy prediction reported as MAE. All methods were tuned with a random search with budget of 500 hyperparameter combinations.}
        \vskip 0.15in
    \begin{small}
    \begin{sc}
    \begin{tabular}{lc}
    \toprule
    {} &           U (internal energy) \\
    \midrule
    RF &  .380 \\
    GCN & .006 \\
    MAT & .004 \\
    \bottomrule
    \end{tabular}
    \end{sc}
    \end{small}
    \label{tab:alchemy}
    \vskip -0.1in
\end{table}

\paragraph{Training error for graph-based neural networks}

\begin{figure}[t!] 
    \centering
    \begin{subfigure}[b]{0.24\textwidth}
        \centering
        \includegraphics[width=\textwidth]{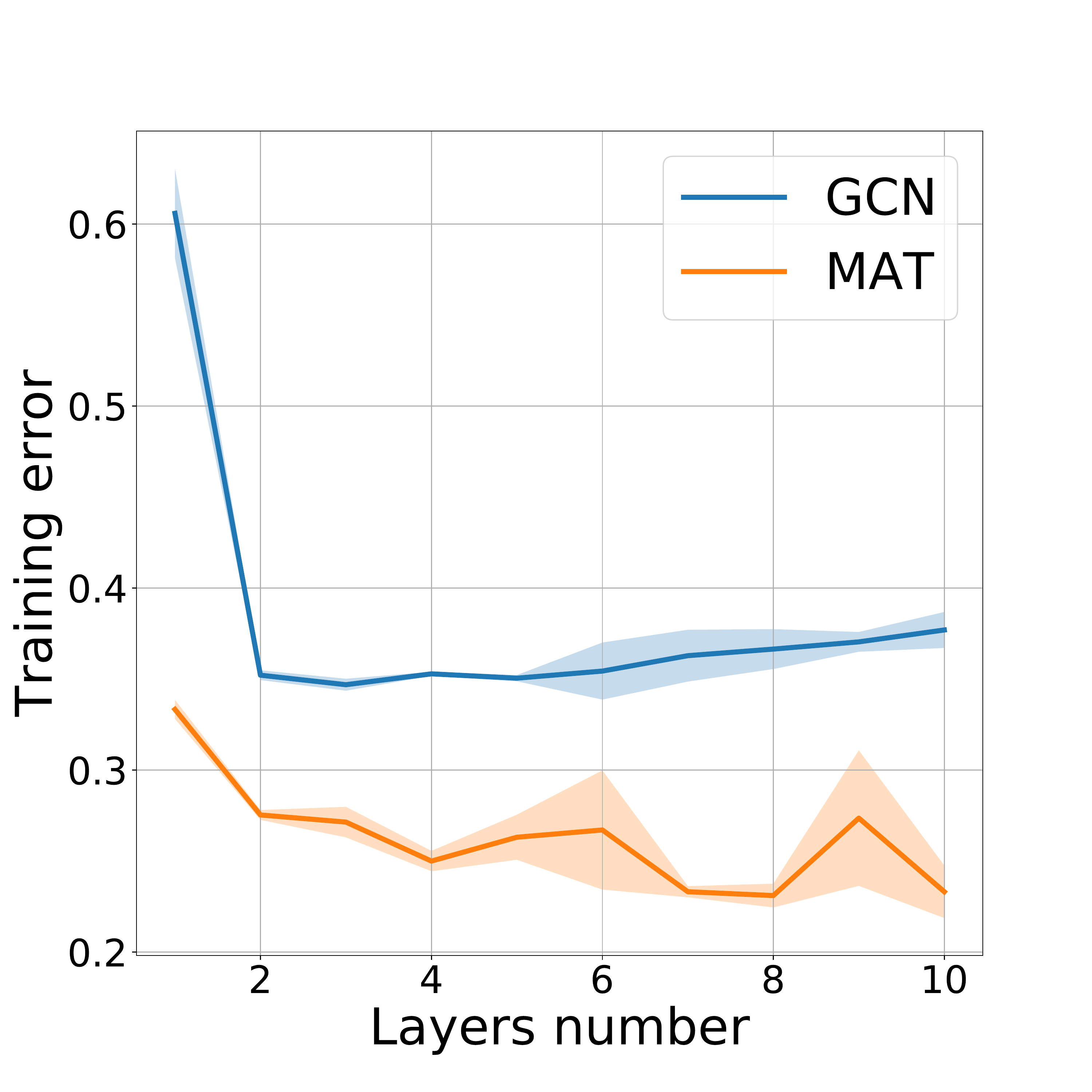}
    \end{subfigure}%
    \begin{subfigure}[b]{0.24\textwidth}
        \centering
        \includegraphics[width=\textwidth]{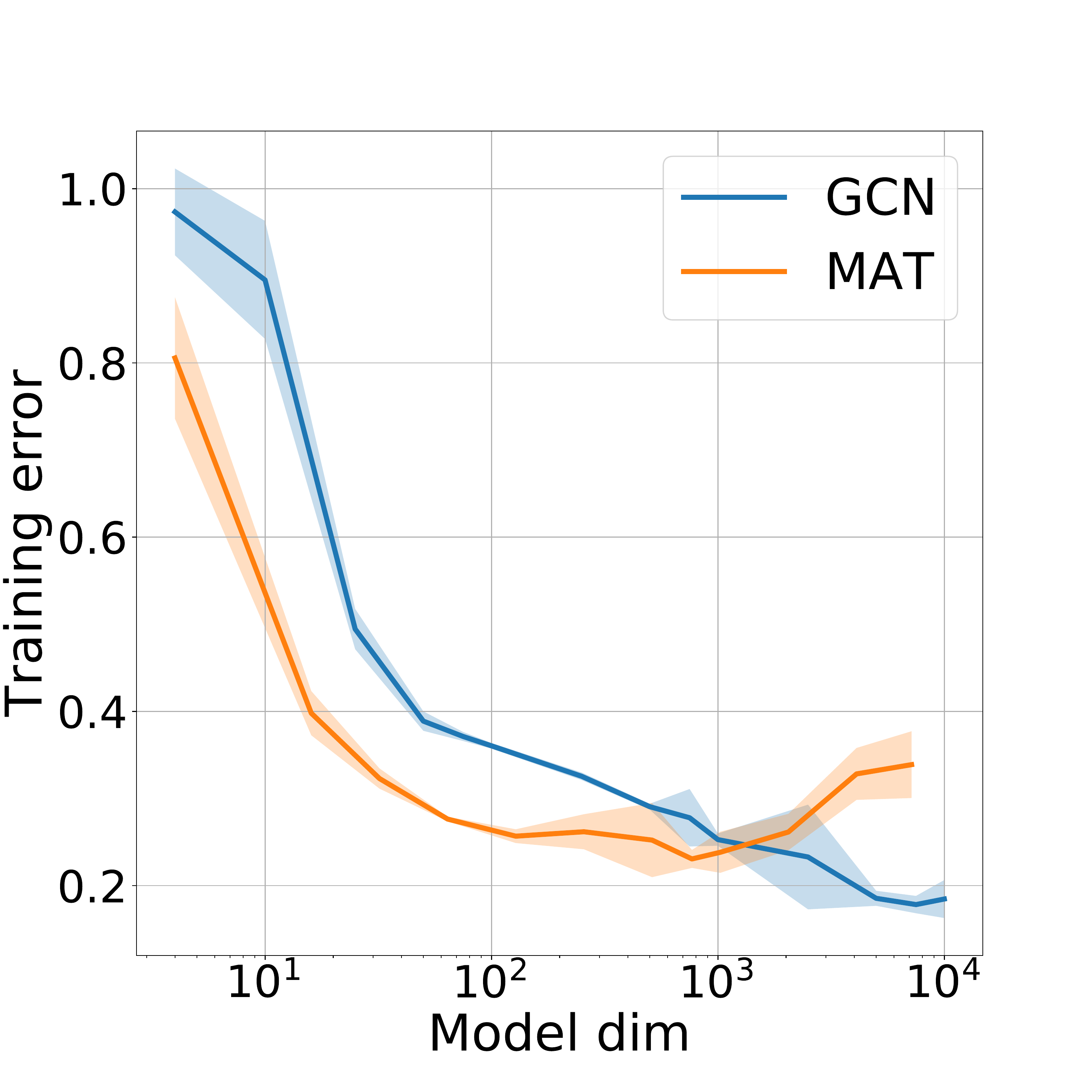}
    \end{subfigure}
    \caption{Training loss of MAT and GCN as a function of the number of layers (left) and model dimensionality (right).}
    \label{fig:gcn_train_loss}
\end{figure}

\citet{Ishiguro2019} show that graph neural networks suffer from underfitting of the training set and their performance does not scale well with the complexity of the network. We reproduce their experiments and confirm that this problem is indeed present for both GCN and MAT. According to Figure \ref{fig:gcn_train_loss}, the training loss of GCN and MAT flattens at some point and stops decreasing even if we keep increasing the number of layers and model dimensionality. Despite this issue, for almost all settings, MAT achieves lower training error than GCN.

\section{Additional details for Sec.~\ref{sec:pretraining}}
\label{app:pretraining}

\paragraph{Task description.} As a node-level pretraining task we chose masking from ~\citep{Hu2019} which is a version of BERT masked language model adapted to graph structured data. The idea is that predicting masked nodes based on their neighbourhood will encourage model to capture domain specific relationships between atoms.

For each molecular graph we randomly replace 15\% of input nodes (atom attributes) with special mask token. After forward pass we apply linear model to corresponding node embeddings to predict masked node attributes. In case of EAGCN we additionally mask attributes of edges connected to masked nodes to prevent model from learning simple value copying.   

\paragraph{Pretraining setting.}
Training dataset consisted of 2 mln molecules sampled from the ZINC15 database. Models were trained for 8 epochs with learning rate set to 0.001 and batch size 256. MAT was optimized with Noam optimizer (described in App. \ref{app:exp_details}), whereas for EAGCN we used Adam \cite{kingma2014adam}. In both cases procedure minimized binary cross entropy loss.

\paragraph{Fine-tuning setting.}
All our pretrained models are fine-tuned on the target tasks for $100$ epochs, with batch size equal to $32$ and learning rate selected from the set of $\{ 1e{-3}, 5e{-4}, 1e{-4}, 5e{-5}, 1e{-5}, 5e{-6}, 1e{-6} \}$. 

In Estrogen Alpha experiments we excluded three molecules (with the highest number of atoms) from the dataset, due to the memory issues.

\begin{table}[h]
\centering
\caption{Pretrained MAT hyperparameters}
\vskip 0.15in
\begin{small}
\begin{sc}
    \begin{tabular}{ll}
    \toprule
    {} &                               parameters \\
    \midrule
    model dim                       &        1024 \\
    layers number                   &           8 \\
    attention heads number          &          16 \\
    PFFs number                     &           1 \\
    $\lambda_{att}$                 &         .33 \\
    $\lambda_{distance}$            &         .33 \\
    distance matrix kernel          &       'exp' \\
    model dropout                   &          .0 \\
    weight decay                    &          .0 \\
    \bottomrule
    \end{tabular}
\end{sc}
\end{small}
\label{app:hp_pretrained_MAT}
\vskip -0.1in
\end{table}

\begin{table}[h]
\centering
\caption{Pretrained EAGCN hyperparameters}
\vskip 0.15in
\begin{small}
\begin{sc}
    \begin{tabular}{ll}
        \toprule
        {} &                    parameters \\
        \midrule
        EAGCN structure &        'weighted' \\
        dropout         &                .0 \\
        weight decay    &                .0 \\
        n conv layers   &                 8 \\
        n dense layers  &                 1 \\
        n sgc         &                1080 \\
        \bottomrule
    \end{tabular}
\end{sc}
\end{small}
\vskip -0.1in
\label{app:hp_pretrained_EAGCN}
\end{table}

\paragraph{SMILES Transformer.}
We used pretrained weights of SMILES-Transformers conducted by \citet{honda2019smiles}. In this setting, according to the authors, we used MLP with $1$ hidden layer, with $100$ hidden units, that works on the $1024$-dimensional molecule embedding returned by the pretrained transformer. We trained this MLP on the target tasks for $100$ epochs, with batch size equal to $32$ and learning rate selected from the set of $\{ 1e{-3}, 5e{-4}, 1e{-4}, 5e{-5}, 1e{-5}, 5e{-6}, 1e{-6} \}$.

\section{Additional results for Sec.~\ref{sec:ablation}}
\label{app:ablation_appendix}

\paragraph{Edge features.}

Every bond in the molecule was embedded by the vector of edge features (we used features similar to described in~\citep{shang2018}). Every edge feature was then passed through linear layer, followed by ReLU activation, which returned one single value for every single edge (if there is no edge between atoms, we pass zero vector through the layer). This results in the matrix  $\mathbf{E} \in \mathbb{R}^{N_{\text{atoms}} \times N_{\text{atoms}}}$ which was then used in Molecule Self-Attention layer, instead of the adjacency matrix.

\begin{table}[h]
\caption{Edge Features used for experiments form Table \ref{tab:ablation_edge}}
\label{tab:edgeatt}
\vskip 0.15in
\begin{small}
\begin{sc}
    \begin{tabular}{ll}
        \toprule
        Attribute & Description \\
        \midrule
        Bond Order & Values from set \{ 1, 1.5, 2, 3 \} \\
        Aromaticity & Is aromatic  \\
        Conjugation & Is conjugated  \\
        Ring Status & Is in a ring  \\
        \bottomrule
    \end{tabular}
\end{sc}
\end{small}
\vskip -0.1in
\end{table}

\section{Toy task}
\label{app:toy}

\paragraph{Task description.} The essential feature of Molecule Attention Transformer is that it augments the self-attention module using molecule structure. Here we investigate MAT on a task heavily reliant on distances between atoms; we are primarily interested in how the performance of MAT depends on $\lambda_a$, $\lambda_d$, $\lambda_g$ that are used to weight the adjacency and the distance matrices in  Equation~\ref{eq:mmselfatt}.

Naturally, many properties of molecules depend on their geometry. For instance, \emph{steric effect} happens when a spatial proximity of a given group, blocks reaction from happening, due to an overlap in electronic groups. However, this type of reasoning can be difficult to learn based only on the graph information, as it does not always reflect the geometry well. Furthermore, focusing on distance information might require selecting low values for either $\lambda_g$ or $\lambda_a$ (see Figure~\ref{fig:architecture}).

\begin{figure}[h]
    \centering
    \begin{subfigure}[b]{0.48\textwidth}
  \centering
  \includegraphics[width=0.7\textwidth]{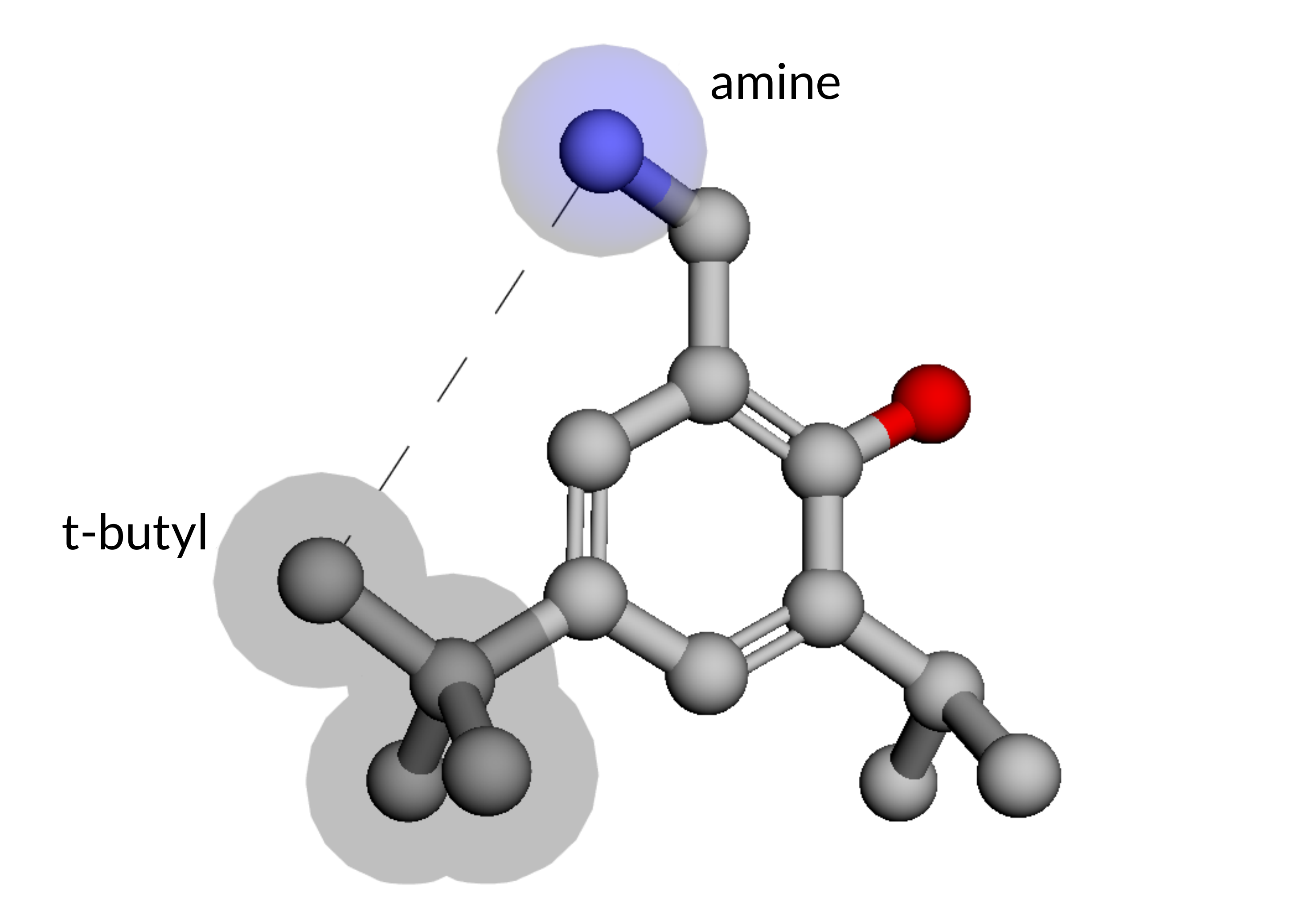}
  \captionof{figure}{The toy task is to predict whether two substructures (-NH$_2$ fragment and \textit{tert}-butyl group) co-occur within given distance.}
  \label{fig:test1}
  \end{subfigure}%
   \hfill
  \raisebox{0.00\height}{\begin{subfigure}[b]{0.48\textwidth}
  \centering
  \includegraphics[width=0.7\textwidth]{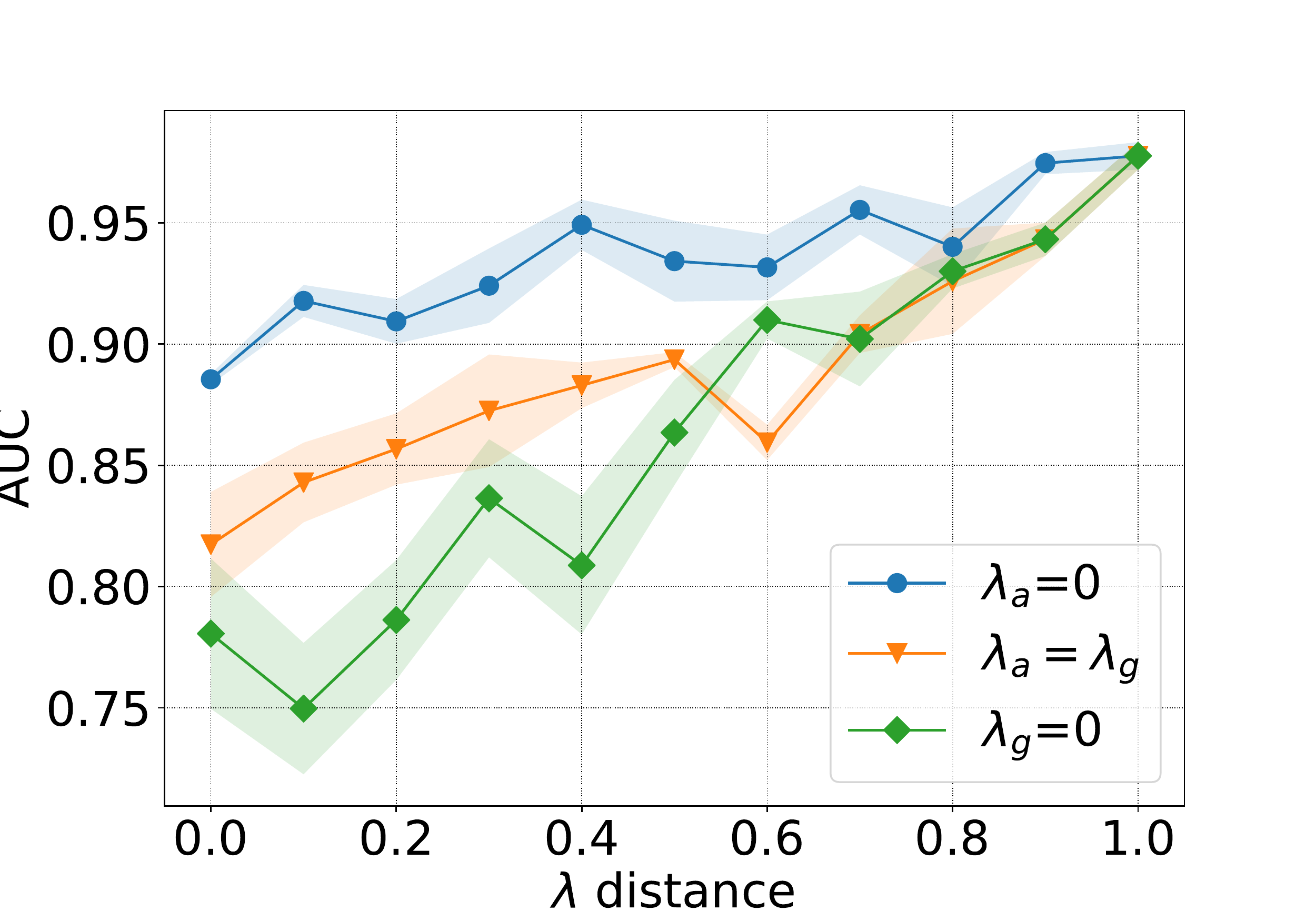}
  \captionof{figure}{Molecule Attention Transformer performance on the toy task as a function of $\lambda_{d}$, for different settings of $\lambda_g$ and $\lambda_a$. }
  \label{fig:toy_task_results}
    \end{subfigure}}
    \caption{MAT can efficiently use the inter-atomic distances to solve the toy task (see left). Additionally, the performance is heavily dependent on $\lambda_d$, which motivates tuning $\mathbf{\lambda}$ in the main experiments (see right).}
    \label{fig:toy_task}
\end{figure}

To illustrate this, we designed a toy task to predict whether or not two substructures are closer to each other in space than a predefined threshold; see also Figure~\ref{fig:test1}. We expect that MAT will work significantly better than a vanilla graph convolutional network if $\lambda_d$ is tuned well.

\paragraph{Experimental setting.} We construct the dataset by sampling 2677 molecules from PubChem~\citep{kim2018pubchem}, and use 20 \AA\, threshold between -NH$_2$ fragment and \textit{tert}-butyl group to determine the binary label. The threshold was selected so that positive and negative examples are well balanced.

\paragraph{Results.} First, we plot Molecule Attention Transformer performance as a function of $\lambda_{d}$ in Figure~\ref{fig:toy_task_results} for three settings of $\mathbf{\lambda}$: $\lambda_a=0$ (blue), $\lambda_a=\lambda_g$ (orange), and $\lambda_g=0$ (green). In all cases we find that using distance information improves the performance significantly. Additionally, we found that GCN achieves $0.93$ AUC on this task, compared to $0.98$ by MAT with $\lambda_{d}=1.0$. These results both motivate tuning $\mathbf{\lambda}$, and show that MAT can efficiently use distance information if it is important for the task at hand.

\paragraph{Further details.} The molecules in the toy task dataset were downloaded from PubChem. The SMARTS query used to find the compounds was \texttt{(C([C;H3])([C;H3])([C;H3]).[NX3H2])}. All molecules were then filtered so that only those with exactly one \textit{tert}-butyl group and one -NH$_2$ fragment were left. For each of them, five conformers were created with RDKit implementation of the Universal Force Field (UFF).

The task is a binary classification of the distance between two molecule fragments. If the euclidean distance between -NH$_2$ fragment and \textit{tert}-butyl group is greater than a given threshold, the label is 1 (0 otherwise). As the distance we mean the distance between the closest heavy atoms in these two fragments across calculated conformers. We used 20 \AA\, as the threshold as it leads to a balanced dataset. There are 2677 compounds total from which 1140 are in a positive class. The dataset was randomly split into training, validation, and test datasets.

In experiments the hyperparameters that yielded promising results on our datasets were used (listed in Table \ref{tab:toy_hyperparameters}). The values of $\lambda$ parameters were tuned, and their scores are shown in Figure~\ref{fig:toy_task_results}. All three $\lambda$ parameters ($\lambda_d$, $\lambda_g$, $\lambda_a$) sum to 1 in all experiments.

To compare our results with a standard graph convolutional neural network, we run a grid search over hyperparameters shown in Table~\ref{tab:gc_toy}. The hyperparameters for which the best validation AUC score was reached are emboldened, and their test AUC score is $0.925 \pm 0.006$.

\begin{table}[h]
    \centering
    \caption{MAT hyperparameters used.}
    
    \vskip 0.15in
    \begin{small}
    \begin{sc}
    \begin{tabular}{rl}
        \toprule
        {}                              &  parameters \\
        \midrule
        batch size                      & 16 \\
        learning rate                   & 0.0005 \\
        epochs                          & 100 \\
        model dim                       & 64 \\
        model N                         & 4 \\
        model h                         & 8 \\
        model N dense                   & 2 \\
        model dense output nonlinearity & 'tanh' \\
        distance matrix kernel          & 'softmax' \\
        model dropout                   & 0.0 \\
        weight decay                    & 0.001 \\
        optimizer                       & 'adam\_anneal' \\
        aggregation type                & 'mean' \\
        \bottomrule
    \end{tabular}
    \end{sc}
    \end{small}
    \label{tab:toy_hyperparameters}
\end{table}
\begin{table}[h]
    \hfill
    \centering
    \caption{Hyperparameters used for tuning GCN.}
    \vskip 0.15in
    \begin{small}
    \begin{sc}
    \begin{tabular}{rl}
        \toprule
        {}                              &  parameters \\
        \midrule
        batch size                      & \textbf{16}, 32, 64 \\
        learning rate                   & \textbf{0.0005} \\
        epochs                          & 20, 40, \textbf{60}, 80, 100 \\
        n filters                       & 64, \textbf{128} \\
        n fully connected nodes         & 128, \textbf{256} \\
        \bottomrule
    \end{tabular}
    \end{sc}
    \end{small}
    \label{tab:gc_toy}

\end{table}

\section{Interpretability analysis}
\label{app:analysis}

\begin{table}[!htb]
    \centering
    \caption{Statistics of the six attention head patterns described in the text. Each head function is defined by a SMARTS that selects atoms with high attention weights. For each atom in the dataset we calculated mean weight assigned to them by the corresponding attention head (average column value of the attention matrix). Calculated means and standard deviations show the difference between attention weights of matching atoms ($\mu^+$, $\sigma^+$) against the other atoms ($\mu^-$, $\sigma^-$).}
    \vskip 0.15in
    \begin{small}
    \begin{sc}
        \begin{tabular}{lcccccc}
            \toprule
            {Head} & i & ii & iii & iv & v & vi \\
            \tiny{SMARTS} & \tiny{[c;D2]} & \tiny{[S,s]} & \tiny{[N;R0]} & \tiny{O=*} & \tiny{[a;D3]} & \tiny{n} \\
            \midrule
            $\mu^+$ & .136 & .330 & .061 & .095 & .043 & .228\\
            $\sigma^+$ & .080 & .280 & .074 & .120 & .032 & .171\\
            $\mu^-$ & .008 & .001 & .002 & .006 & .006 & .005\\
            $\sigma^-$ & .032 & .003 & .016 & .034 & .014 & .009\\
            \bottomrule
        \end{tabular}
     \label{tab:heads_overall}
     \end{sc}
     \end{small}
\end{table}

We found several patterns in the self-attention heads by looking at the first layer of MAT. These patterns correspond to chemical structures that can be found in molecules. For each such pattern found in a qualitative manner, we tested quantitatively if our hypotheses are true about what these particular attention heads represent.

For each pattern found in one of the attention heads, we construct a SMARTS expression describing atoms that belong to our hypothetical molecular structures. Then, all atoms matching the pattern are extracted from the BBBP dataset, and their mean attention weights (average column value of the attention matrix) are compared against atoms that do not match the pattern. Table~\ref{tab:heads_overall} shows the distributions of attention weights for matching and not matching atoms. Atoms which match the SMARTS expression have significantly higher attention weights ($\mu^+ > \mu^-$).


\end{document}